\documentclass{article}
\pdfoutput=1

\usepackage{arxiv}

\usepackage[utf8]{inputenc} % allow utf-8 input
\usepackage[T1]{fontenc}    % use 8-bit T1 fonts
\usepackage{hyperref}       % hyperlinks
\usepackage{url}            % simple URL typesetting
\usepackage{booktabs}       % professional-quality tables
\usepackage{amsfonts}       % blackboard math symbols
\usepackage{nicefrac}       % compact symbols for 1/2, etc.
\usepackage{microtype}      % microtypography
\usepackage{lipsum}		% Can be removed after putting your text content
\usepackage{graphicx}
\usepackage[numbers]{natbib}
\usepackage{doi}

\usepackage{subcaption}
\usepackage{amsmath}
\usepackage{algorithm}
\usepackage{algpseudocode}

\title{Physics-Informed Neural Networks for High-Frequency and Multi-Scale Problems using Transfer Learning}

\author{\href{https://orcid.org/0009-0000-9337-7410}{\includegraphics[scale=0.06]{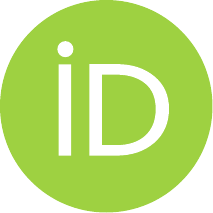}\hspace{1mm}Abdul Hannan Mustajab} \thanks{These authors contributed equally to this work.}\\
	DIBRIS\\
 University of Genoa\\
 via Dodecaneso, 35, 16146 Genoa, Italy\\
	\texttt{S5156186@studenti.unige.it} \\ 0009-0000-9337-7410
\And
	\href{https://orcid.org/0000-0002-9107-2880}{\includegraphics[scale=0.06]{orcid.pdf}\hspace{1mm}Hao Lyu}\thanks{These authors contributed equally to this work.} \\
	Institute for Geosciences\\Kiel University\\
       24118 Kiel, Germany\\
	\texttt{hao.lyu@ifg.uni-kiel.de} \\
 \And
	\href{https://orcid.org/0000-0002-7553-8310}{\includegraphics[scale=0.06]{orcid.pdf}\hspace{1mm}Zarghaam Rizvi} \thanks{Current address: GeoAnalysis Engineering GmbH, Kiel, Germany 24118.}\\
 Department of Civil and Environmental Engineering\\
	University of Waterloo\\
 Waterloo, Canada N2L 3G1\\
	\texttt{zarghaam.rizvi@geoanalysis-engineering.de} \\
 \And
	\hspace{1mm}Frank Wuttke \thanks{Corresponding author.}\\
	Institute for Geosciences\\Kiel University\\
       24118 Kiel, Germany\\
	\texttt{frank.wuttke@ifg.uni-kiel.de} \\
}

\renewcommand{\shorttitle}{\textit{arXiv} Template}

\hypersetup{
pdftitle={A template for the arxiv style},
pdfsubject={q-bio.NC, q-bio.QM},
pdfauthor={David S.~Hippocampus, Elias D.~Striatum},
pdfkeywords={First keyword, Second keyword, More},
}

\begin{document}
\maketitle

\begin{abstract}
  Physics-informed neural network (PINN) is a data-driven solver for partial and ordinary differential equations(ODEs/PDEs). It provides a unified framework to address both forward and inverse problems. However, the complexity of the objective function often leads to training failures. This issue is particularly prominent when solving high-frequency and multi-scale problems. We proposed using transfer learning to boost the robustness and convergence of training PINN, starting training from low-frequency problems and gradually approaching high-frequency problems. Through two case studies, we discovered that transfer learning can effectively train PINN to approximate solutions from low-frequency problems to high-frequency problems without increasing network parameters. Furthermore, it requires fewer data points and less training time. We elaborately described our training strategy, including optimizer selection, and suggested guidelines for using transfer learning to train neural networks for solving more complex problems.
\end{abstract}

% keywords can be removed

\section{Introduction}
Physics Informed Neural Networks (PINNs) are a relatively new data-driven solver of partial differential equations (PDEs) \cite{raissi2017physics, Raissi2017PhysicsID, RAISSI2019686}. The neural networks' capability to approximate complex functions is their basis for solving partial differential equations. While the idea of using neural networks to estimate PDE solutions dates back to the 1990s, it initially garnered limited attention for various reasons. With the rapid advancements in deep neural network technology, the exponential growth in computing power, and the thriving open deep learning community, PINNs have recently garnered substantial interest and acclaim.

PINNs possess several notable advantages that make them a competitive method compared to mature, traditional numerical approaches for PDEs. PINN, as a meshless method, directly embeds mathematical equations into the network structure. The dual reliance on observational data and mathematical models equips PINN to handle noisy observational data. Moreover, PINN offers a consistent framework for forward and inverse problems through optimization algorithms \cite{RAISSI2019686}. By simply extending the neural network with additional output channels, PINNs can be employed to solve inverse problems. In inverse design, PINNs can impose PDEs as rigorous constraints, enhancing their utility. While neural networks grapple with the curse of dimensionality as problems become more complex, PINNs strive to resolve PDEs and their inversion challenges in domains characterized by intricate geometries and high dimensions, where numerical simulations are notably challenging.

PINNs harness the PDEs to guide and constrain the training process of neural networks. PINNs incorporate the residuals, initial conditions, and boundary conditions of the PDE into their loss function. The neural networks are tasked with fitting the observed data while simultaneously minimizing the PDE residuals. The neural network is thus trained to approximate the solution of the PDE by minimizing this loss function. This approach reduces the need for additional observational data, making it a powerful and efficient technique to solve PDEs.

The core of PINN implementation is to calculate partial derivatives, and this task can be completed through automatic differentiation algorithms in mainstream deep learning frameworks \cite{baydin2018automatic}. Several open-source libraries such as DeepXDE \cite{doi:10.1137/19M1274067}, SimNet \cite{10.1007/978-3-030-77977-1_36} and SciANN \cite{HAGHIGHAT2021113552} have been developed, making PINN easier to apply in practice. PINN has produced compelling results on a range of problems in computational science and engineering, such as computational fluid dynamics, acoustics, solid mechanics \cite{LIANG2023108575, doi:10.1061/(ASCE)EM.1943-7889.0001947, math11112529}, and geo-physics\cite{Cai2021, WANG2023106872, HAGHIGHAT2021113741, Okazaki2022, doi.org/10.1029/2021JB023120, KARIMPOULI20201993}.

The challenge of training PINNs to achieve fast convergence and accuracy is persistent. This challenge is intricately linked to the highly complex and non-convexity of the loss function, which makes a PINN hard to train\cite{NEURIPS2021_df438e52, doi:10.1137/20M1318043}. Besides, training PINNs also suffer from spectral bias. The neural networks prioritize learning low-frequency patterns over high-frequency details \cite{pmlr-v97-rahaman19a, 7f26e73e6f7545e6bc395a7caf1b8c70}. When the problem contains high-frequency features, the PINN models often fail to converge to the desired solution due to this phenomenon \cite{9903391, WANG2021113938, NEURIPS2020_55053683}. PINNs inherent ability to encapsulate domain knowledge and exploit neural network architectures has made them particularly attractive for simulating complex physical systems. However, as the applications of PINNs extend to problems characterized by high-frequency oscillations and intricate multiscale phenomena, they face significant hurdles. These challenges often manifest in numerical instability, slow convergence, and increased computational demands, making it imperative to develop strategies that enhance the robustness and efficiency of PINNs in such scenarios.

We believe that transfer learning is a key technique to address the training difficulties for high-frequency and multiscale problems. Transfer learning is a technique that leverages knowledge acquired from solving one problem to solve a related problem, involving training a network to solve the desired PDEs from an initial model \cite{Markidis1621250}. It enables training PINNs with a reduced amount of data and training costs \cite{Prantikos2023, XU2023115852,TANG2022113101}. Moreover, it addresses the challenge of insufficient high-fidelity data in numerous scientific computing cases \cite{CHAKRABORTY2021109942}. Through transfer learning, PINN demonstrates its capability to effectively solve intricate PDEs, positioning itself as a valuable tool in addressing complex engineering challenges, such as fracture mechanics\cite{GOSWAMI2020102447} and flows in porous media\cite{chen2023transfer}.

The primary objective of this research is to elucidate the prevailing challenges encountered in PINNs when applied to high-frequency and multiscale problems. To mitigate these challenges, we investigate the utility of transfer learning. By incorporating transfer learning into the PINN framework, we aim to harness the benefits of pre-trained models and transferable knowledge, potentially enhancing the convergence and accuracy of PINNs for high-frequency and multiscale applications. Moreover, the choice of optimizer plays a crucial role in training neural networks, including PINNs. Different optimization algorithms possess distinct characteristics and may perform differently in terms of convergence speed and solution quality. In this study, we empirically evaluate a range of optimizers to determine their effectiveness in training the foundational model of the PINN. Through a comparative analysis, we seek to identify the optimizer that best suits the specific requirements and challenges of PINNs in the context of high-frequency and multiscale problems. We take wave propagation for our case study as it is an essential phenomenon in engineering due to its ability to transfer energy and information through a medium without the bulk motion of the medium itself. Waves are a fundamental concept in many engineering disciplines, including acoustics, electro-magnetics, cosmology, fluid dynamics, and not least in geophysics \cite{Okazaki2022, doi.org/10.1029/2021RG000742, moseley2020solving}. In particular, PINN has been explored and applied to full waveform inversion (FWI) due to its ability to solve inverse problems with noisy inputs \cite{doi.org/10.1029/2021JB023120, kollmannsberger2023transfer, doi.org/10.1029/2022JB025493, 10035474, 10.1093/gji/ggad215}. The wave equation provides a mathematical framework for understanding and predicting how waves propagate through various physical systems. While numerous numerical methods have been developed for solving wave equations, the emergence of PINN has garnered significant interest as a data-driven approach \cite{10096980, s23052792, NGUYEN2023111828, 10.1007/978-3-031-35995-8_6}. 

In summary, this manuscript addresses these issues faced by PINNs when confronted with high-frequency and multiscale problems. By investigating transfer learning and scrutinizing the performance of various optimizers, we aim to provide valuable insights into improving the efficacy and versatility of PINNs for challenging physical simulations. The rest of the paper is organized as follows: The second section provides a brief introduction to PINN, emphasizing the crucial components pertinent to our study. In the third section, we show two studies where transfer learning is employed to train PINN for solving partial differential equations (PDEs) from low frequencies to high frequencies. Additionally, we explore best practices for selecting the base model. The final section summarizes our findings and offers conclusions while also suggesting potential directions for future research.

\section{Method}
\subsection{Physics Informed Neural Networks}
PINNs or Physics-Informed Neural Networks are a specific kind of neural network that is trained to approximate the solution to any given law of physics defined by a partial differential equation (PDE) or a system of PDEs \cite{RAISSI2019686}. The most significant benefit of PINN over other methods is that it is a mesh-free method. The classical PINN follows the collocation-based approach, implying that the neural network aims to approximate the strong form of the governing equation at a set of collocation points. As the collocation points can be distributed randomly within the domain, and no mesh is required, this approach belongs to the category of mesh-free methods \cite{Anitescu2023}. Most modern machine learning frameworks, such as Pytorch or Tensorflow, have implemented automatic differentiation for PINNs.

The architecture of a PINN can vary depending on the specific problem, while many PINNs still use the feed-forward fully connected neural network (FCN) as part of their architecture. The FCN is the basic architecture used in deep learning algorithms~\cite{lecun2015deep}. A fully-connected neural network with $L$ layers is a function $f_{\theta} : \mathbb{R}^d \to \mathbb{R}^k$ described by

\begin{equation}
f_{\theta}(x) = W^{[L-1]}\sigma \circ (\ldots \sigma \circ (W^{[0]}x+b^{[0]}) + \ldots) + b^{[L-1]},
\label{eq:nn_function}
\end{equation}

where $\sigma$ is an entry-wise activation function, $W^{[l]}$ and $b^{[l]}$ are respectively the weight matrices and the bias corresponding to each layer $l$, and $\theta$ is the set of weights and biases:

\begin{equation}
\theta = (W^{[0]}, \ldots , W^{[L-1]}, b^{[0]}, \ldots , b^{[L-1]}).
\label{eq:nn_parameters}
\end{equation}

The activation function is a crucial component of a neural network, and there are several favoured choices available, including the sigmoid function, hyperbolic tangent function (tanh), and rectified linear unit (ReLU). It is worth mentioning that we have implemented the hyperbolic tangent function as part of the neural network for PINN.

The hyperbolic tangent activation function is defined as 
\[
tanh(x) = \frac{e^{x} - e^{-x}}{e^{x} + e^{-x}}.
\]

% \begin{figure}[H]
% \centering
%   \includegraphics[width=1\linewidth]{definitions/tanh.png}
%   \caption{Hyperbolic Tangent Function}
%   \label{tanh_figure}
% \end{figure}

The smoothness and overall S-shape of this function are similar to that of the sigmoid function. However, unlike the sigmoid function, the range of the outputs is centered at 0 and falls between (-1, 1). This makes the tanh activation function more appropriate for deep neural networks as it avoids creating a bias towards positive outputs\cite{Anitescu2023}. ReLU is more commonly used as an activation function in neural networks. However, it's unsuitable for PINNs due to its second derivative being zero.

\include{methodology}.

\begin{figure}[H]
\centering
  \includegraphics[width=0.75\linewidth]{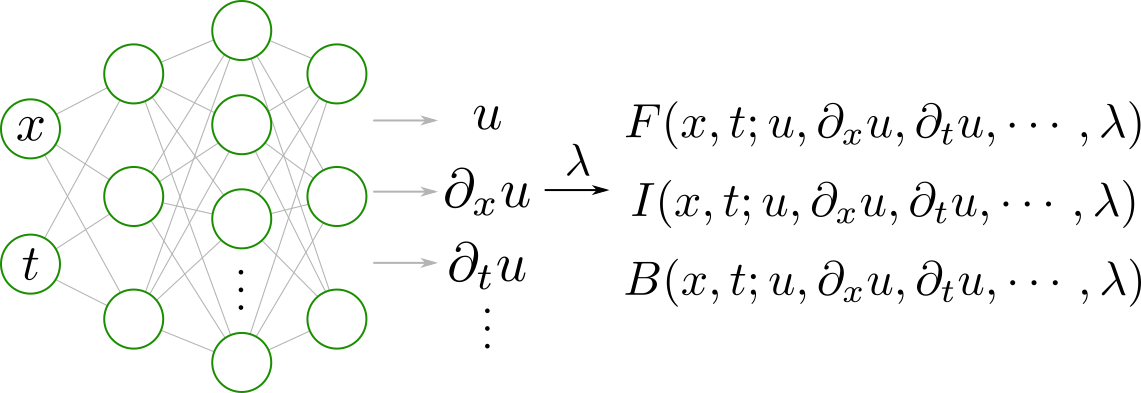}
  \caption{PINN Model}
  \label{fig_pinn}
\end{figure}

A PINN consists of multiple loss terms, each corresponding to  initial conditions, boundary conditions and the function loss itself, or the PDE / ODE residuals. This results in a high-dimensional and non-convex loss function with different competing loss terms. It is essential to weigh these loss terms; otherwise, the optimizer might train only one term and create a bias. Later in the 1D wave section, we discuss about temporal loss weighting technique that we used to assign the highest weight to temporal loss terms in the beginning. 

We consider a scalar function $u(x,t)$ on the domain $\Lambda \times [0, \infty)$;  with the boundary $\partial \Lambda$, where $\Lambda \subset \mathbb{R}^d$. $u(x,t)$ satisfies the following PDEs:

\begin{align}
F(x,t;u,\partial_x u,\partial_t u, \ldots, \lambda) &= 0, \quad \forall (x,t) \in U \\
I(x,t_0,h;u,\partial_t u, \ldots, \lambda) &= 0, \quad \forall (x,t) \in \mathcal{I} \\
B(x,t,g;u,\partial_x u, \ldots, \lambda) &= 0, \quad \forall (x,t) \in \partial U
\end{align}

where $F$ contains a sequence of differential operators (i.e., $[\partial_t, \partial_x, \ldots]$), which represent the residual of the PDEs, $\lambda$ is the PDEs' parameter vector, $I$ is the residual form of the initial condition containing a function $h(x,t)$, and $B$ is the residual form of the boundary condition containing a function $g(x,t)$. $U = \{(x,t) \,|\, x \in \Lambda, t \in [0, T]\}$, $\partial U = \{(x,t) \,|\, x \in \partial \Lambda, t \in [0, T]\}$, and $\mathcal{I} = \{(x,t) \,|\, x \in \partial \Lambda, t = 0\}$.

Figure \ref{fig_pinn} illustrates the structure of a PINN model. The space coordinates $x$ and time $t$ are usually taken as the inputs, and the outputs $\hat{u}(x,t)$ are used to approximate the true solution $u(x,t)$ of the PDEs. The differential operators are calculated by Automatic Differentiation (AD), and then the PDEs' residual, initial condition and boundary condition are embedded into the loss function of neural networks:

\begin{equation}
L(\theta) = W_F \mathcal{L}_F(\theta) + W_I \mathcal{L}_I(\theta) + W_B \mathcal{L}_B(\theta)
\end{equation}

With $\theta$ representing the weights of neural network, $W_F$, $W_I$, and $W_B$ are the weights for various loss terms, and $\mathcal{L}_F$, $\mathcal{L}_I$, and $\mathcal{L}_B$ are the loss functions of PDE, initial condition, and boundary condition, respectively:

\begin{align}
\mathcal{L}_F &= \frac{1}{N_F} \sum_{i=1}^{N_F} \|F(x^{(i)}, t^{(i)}; \hat{u})\|^2 \\
\mathcal{L}_I &= \frac{1}{N_I} \sum_{i=1}^{N_I} \|I(x^{(i)}, t^{(i)}, h^{(i)}; \hat{u})\|^2 \\
\mathcal{L}_B &= \frac{1}{N_B} \sum_{i=1}^{N_B} \|B(x^{(i)}, t^{(i)}, g^{(i)}; \hat{u})\|^2
\end{align}

where $N_F$, $N_I$, and $N_B$ are the sets of collocation points in $U$, $\mathcal{I}$, and $\partial U$, and $N_F$, $N_I$, and $N_B$ denote the number of sampling points. In this manuscript, the total loss function is represented as $L_{\text{PINN}}(\theta)$.

After the formulation of these loss terms, the PINN can be trained using any optimizer, such as Adam, Stochastic Gradient descent or Netwon-based method like L-BFGS. In this work, we mainly use Adam \cite{kingma2014adam} and LBFGS \cite{BFGS}, which are described in detail in the following parts.
\subsection{Optimizers}

Here we outline the common optimization algorithms used to train neural networks, and minimizing the loss function.

\subsubsection{Adam Optimizer}

\begin{algorithm}
    \caption{Adam Optimization}
    \begin{algorithmic}
        \State \textbf{Input:} parameters, learning\_rate, $\beta_1$, $\beta_2$, $\epsilon$
        \State $m \gets \text{zeros\_like}(parameters)$  \Comment{Initialize 1st moment vector}
        \State $v \gets \text{zeros\_like}(parameters)$  \Comment{Initialize 2nd moment vector}
        \State $t \gets 0$  \Comment{Initialize timestep}
        
        \While{not converged}
            \State $t \gets t + 1$
            \State $\text{gradient} \gets \text{compute\_gradient}(parameters)$  \Comment{Compute gradient of the objective function}
            \State $m \gets \beta_1 \cdot m + (1 - \beta_1) \cdot \text{gradient}$  \Comment{Update biased first moment estimate}
            \State $v \gets \beta_2 \cdot v + (1 - \beta_2) \cdot (\text{gradient}^2)$  \Comment{Update biased second raw moment estimate}
            \State $m_{\text{hat}} \gets m / (1 - \beta_1^t)$  \Comment{Bias-corrected first moment estimate}
            \State $v_{\text{hat}} \gets v / (1 - \beta_2^t)$  \Comment{Bias-corrected second moment estimate}
            \State $parameters \gets parameters - \text{learning\_rate} \cdot m_{\text{hat}} / (\sqrt{v_{\text{hat}}} + \epsilon)$  \Comment{Update parameters}
        \EndWhile
    \end{algorithmic}
\end{algorithm}

\subsubsection{LBFGS Optimizer}

Broyden–Fletcher–Goldfarb–Shanno is a quasi-Newton-based optimization algorithm commonly used for training neural networks. The loss landscape of a PINN is highly complex due to competing loss terms, making BFGS an effective choice for training PINNs.

BFGS (\cite{bfgs1}) is a gradient method that iteratively computes the Hessian matrix of the loss function, and this process requires $O(n^2)$ gradient evaluations, where n represents the number of parameters. The BFGS curvature matrix can be updated without the need for matrix inversion, and this reduces the computational cost significantly. However, since the Hessian matrix is the foundation of the BFGS algorithm, memory usage increases as the square of the number of parameters. This results in rapid memory usage growth, making it impractical to use this approach for neural networks with a large number of parameters.

The BFGS algorithm may use large amounts of memory, but L-BFGS (\cite{lbfgs}) solves this issue by storing a few vectors that represent an estimate of the full Hessian matrix. Compared to BFGS, L-BFGS is more computationally efficient, uses less memory, and can handle problems with larger numbers of parameters. Due to its lower memory requirements, the L-BFGS algorithm has become the favorite among second-order optimization techniques.

\begin{algorithm}
    \caption{BFGS Method}
    \begin{algorithmic}[1]
        \State \textbf{Input:} Initial guesses $x_0$ and $B_0$, tolerance \text{tol}
        \State Set $k = 0$
        \Repeat
            \State Obtain descent direction $d_k = -B_k^{-1}\nabla f(x_k)$
            \State Set $\alpha_k = 1$
            \State Calculate the step $s_k = \alpha_k d_k$
            \State Update the design $x_{k+1} = x_k + s_k$
            \If{$|x_{k+1} - x_k| < \text{tol}$ or $|\nabla f(x_{k+1})| < \text{tol}$}
                \State \textbf{break}
            \EndIf
            \State Obtain the variation in the gradient $y_k = \nabla f(x_{k+1}) - \nabla f(x_k)$
            \State Update the Hessian approximation $B_{k+1} = B_k$
            \State Increase the iterator $k = k + 1$
        \Until{convergence}
    \end{algorithmic}
\end{algorithm}

\section{Result and discussion}
\subsection{Simple Harmonic Oscillator (SHM).}
The damped harmonic oscillator is a classic problem in mechanics that describes the motion of a mechanical oscillator (e.g., a spring pendulum) under the influence of a restoring force and friction. The governing equation for the damped harmonic oscillator is given by:

\begin{equation}
    F(x) = m\frac{d^2u}{dt^2} + \mu\frac{du}{dt} + ku
\end{equation}

where:
\begin{align*}
    m & : \text{mass of the oscillator} \\
    \mu & : \text{coefficient of friction} \\
    k & : \text{spring constant}
\end{align*}

In the paper, we focus on the under-damped state, i.e. where the oscillation is slowly damped by friction occurs when $\delta < \omega_0$, where $\delta = \frac{\mu}{2m}$ and $\omega_0 = \sqrt{\frac{k}{m}}$.

The following initial conditions are applied:
\[
u(t=0) = 1, \quad \frac{du}{dt}(t=0) = 0
\]

The exact solution of the above setup is given by:
\begin{equation}
u(t) = e^{-\delta t} \left(2A \cos(\phi + \omega t)\right), 
\end{equation}
where $\omega = \sqrt{\omega_0^2 - \delta^2}$.

The interior residual is given by
\begin{equation}
r_{\text{int},\theta}(t):= m\frac{d^2\hat{u}_\theta}{dt^2} + \mu\frac{d\hat{u}_\theta}{dt} + k\hat{u}_\theta
\end{equation}

\begin{center}
This is the exact solution of the oscillator with $w_0$ as $20$ Hz.
\end{center}
% \begin{figure}[H]
% \includegraphics[width=10.5cm]{definitions/actual-solution.png}
% \end{figure}

\begin{figure}[H]
    \centering
    \includegraphics[width=0.18\textwidth]{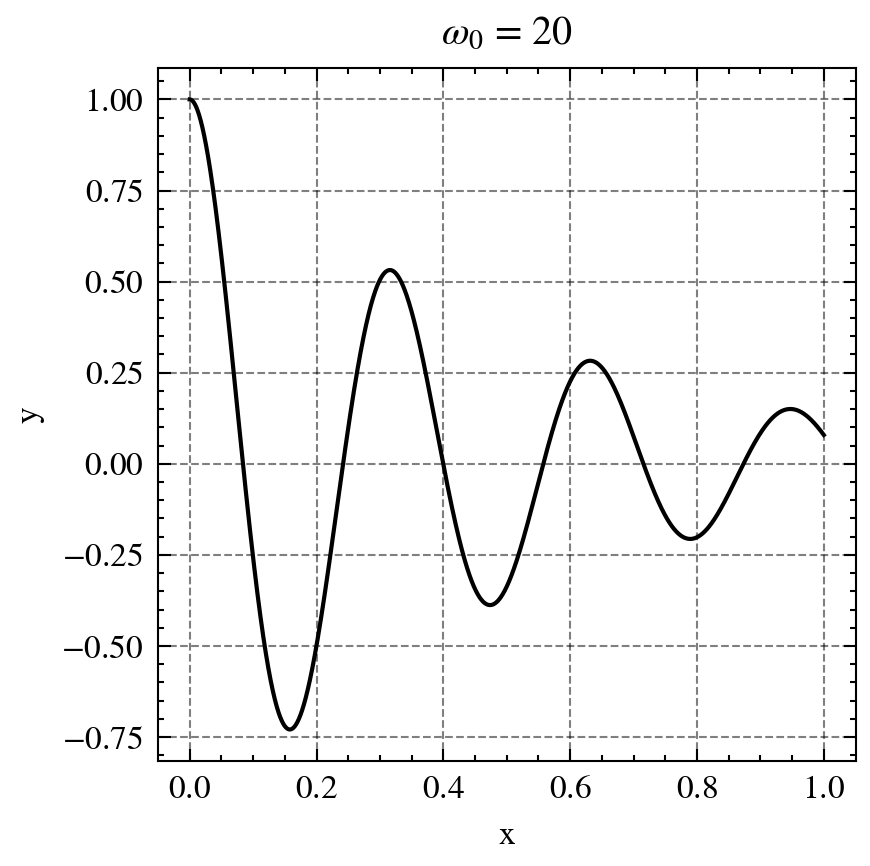}
    \includegraphics[width=0.18\textwidth]{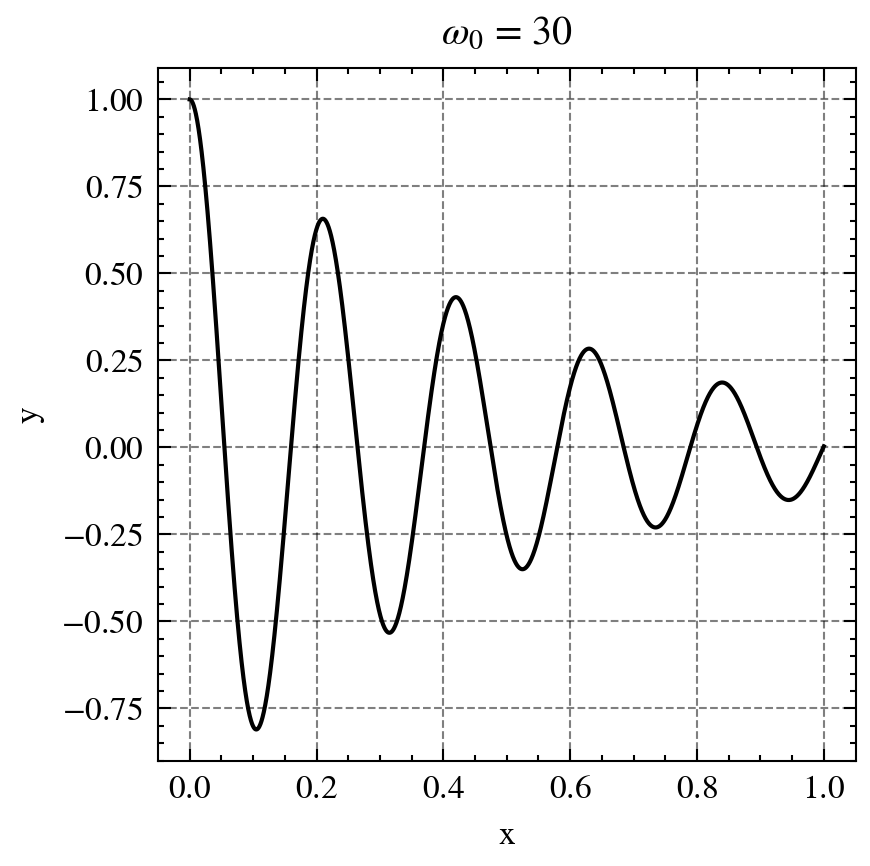}
    \includegraphics[width=0.18\textwidth]{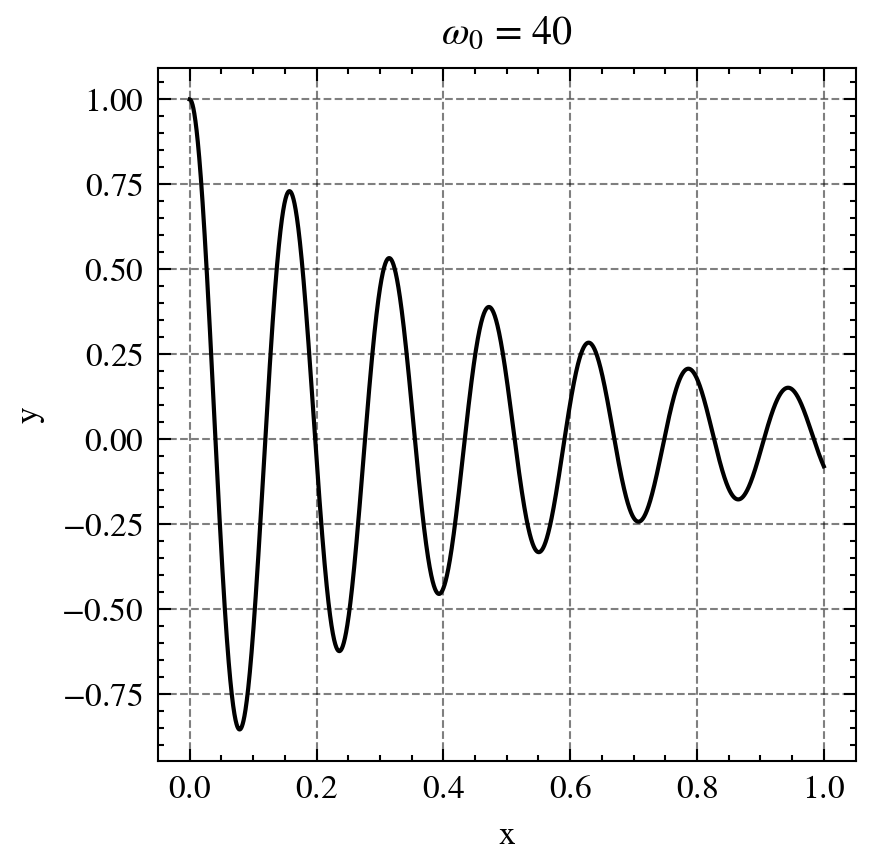}
    \includegraphics[width=0.18\textwidth]{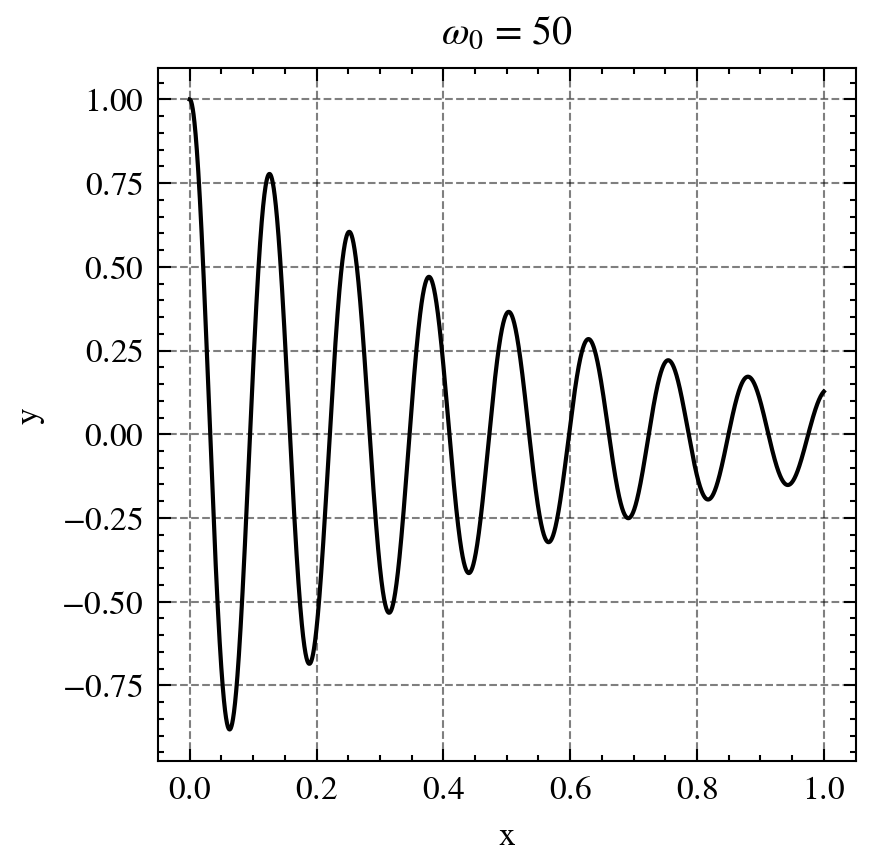}
    \includegraphics[width=0.18\textwidth]{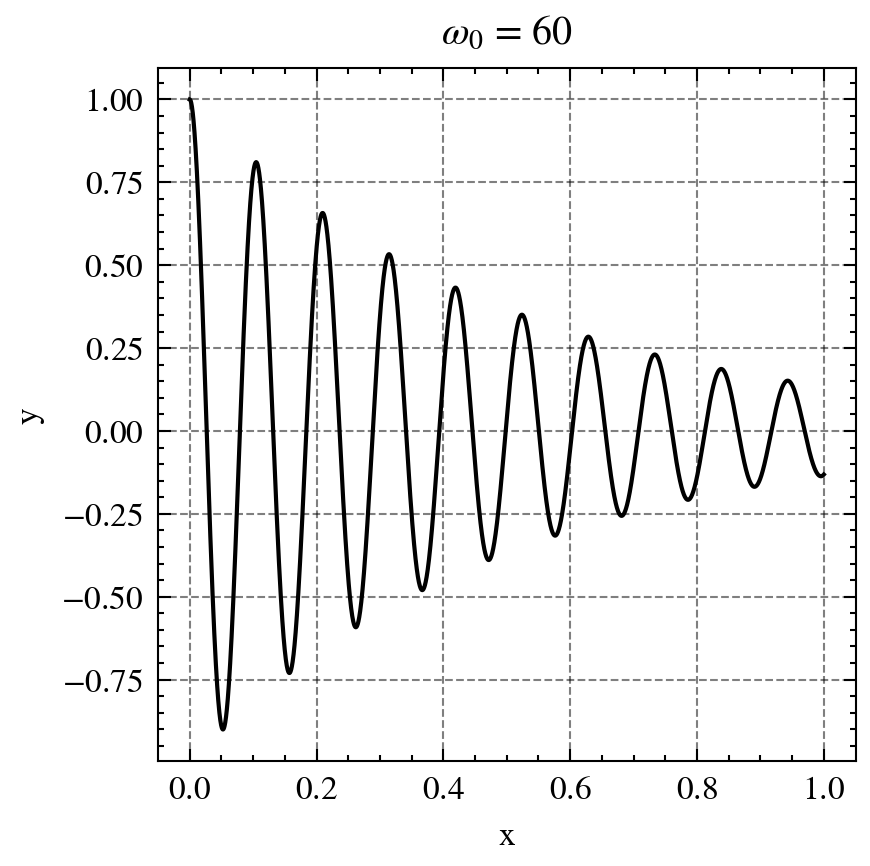}
    \centering
    \caption{Exact solution for different values of $\omega$}
    \label{figure:shm_solution}
\end{figure}

With an increasing frequency ($\omega_0$), the damped harmonic oscillator function becomes more complicated for PINNs to approach. Figure \ref{figure:shm_solution} illustrates the exact solution of the oscillator for four frequencies $\omega_0 = 20, 40, 50, 60$. 

In the experiment, our PINN model is used to approximate the solutions of the oscillator for the above four frequencies. The selected source terms yield uncomplicated solutions that demonstrate how the F-principle affects the convergence of PINN to the numerical solution. According to the F-principle, the low-frequency or large-scale characteristics of the solution are initially manifested in the PINNs, while it may take multiple training epochs to retrieve the high-frequency or small-scale features.\cite{Markidis1621250}. We expect that the vanilla PINN will converge faster and achieve better accuracy in learning the damped harmonic oscillator for lower-frequency components, e.g., $\omega_0 = 20$ than for higher-frequency components ($\omega_0 = 40, 50, 60$). The experiment results that come in the following part are aligned with the expectations.

The PINN model we used in experiments for this case, comprises a fully connected network (FCN) with 5 fully connected layers, each consisting of 64 neurons, totalling 4321 parameters. We trained the PINN model using two optimizers, Adam and L-BFGS, which are mentioned in most PINN papers. To populate the computational domain, we utilized a total of 100 equidistant points. It is worth noting that the selection of the number of points within the domain is a decision that is dependent on the user.

%We will use two different approaches to solve this problem. One will be using PINN with Adam Optimizer, and the second will be to use transfer learning to use the weights learned by Adam PINN and further train it using L-BFGS.

%$w_0 = 20, 40, 60, 80$

\begin{figure}[H]
  \centering
  \begin{subfigure}{0.45\textwidth}
    \centering
    \includegraphics[width=\linewidth]{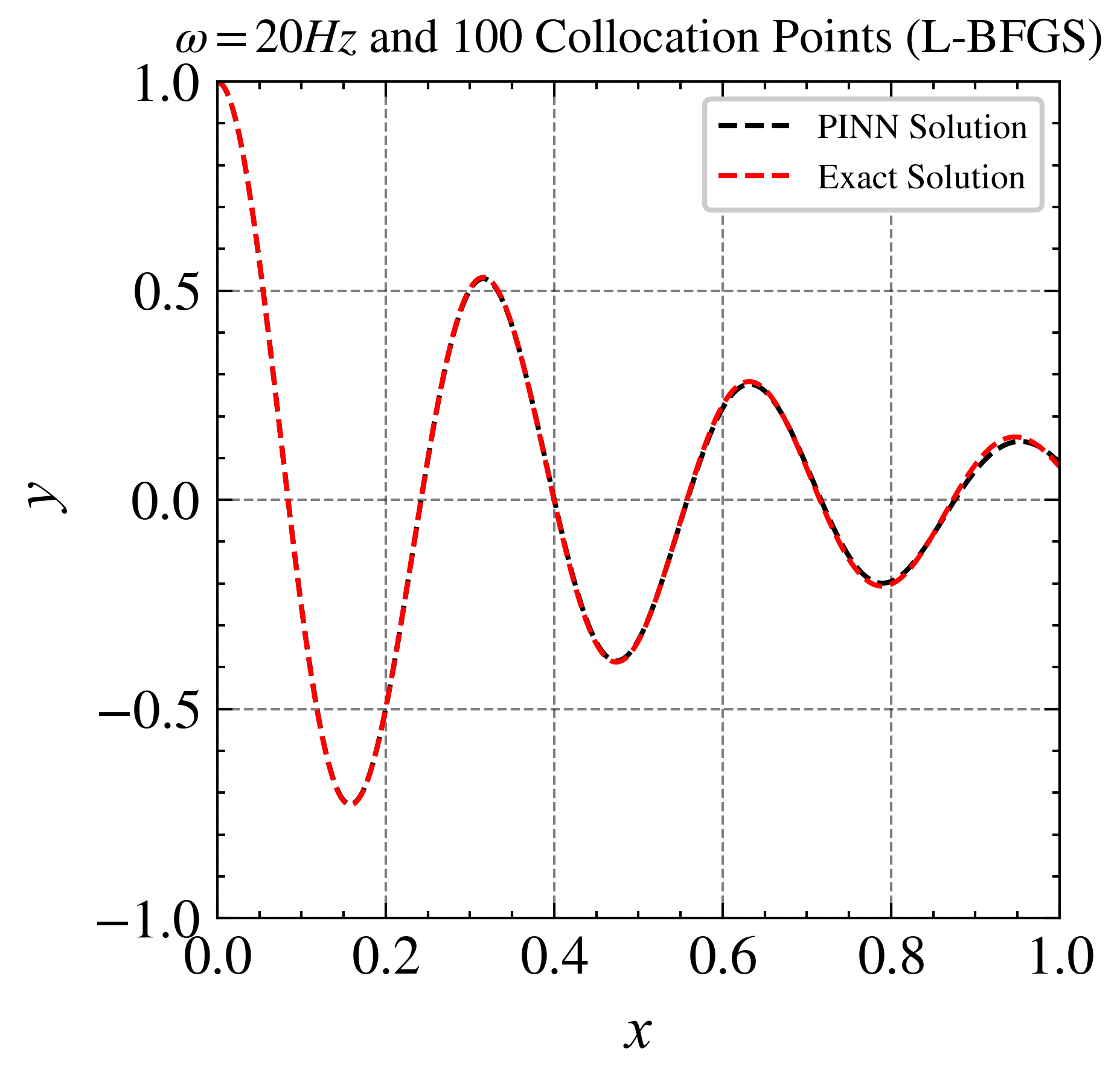}
    \caption{PINN solution vs. exact solution using L-BFGS}
    \label{fig:20hz_pinn_a}
  \end{subfigure}
  \hfill
  \begin{subfigure}{0.45\textwidth}
    \centering
    \includegraphics[width=\linewidth]{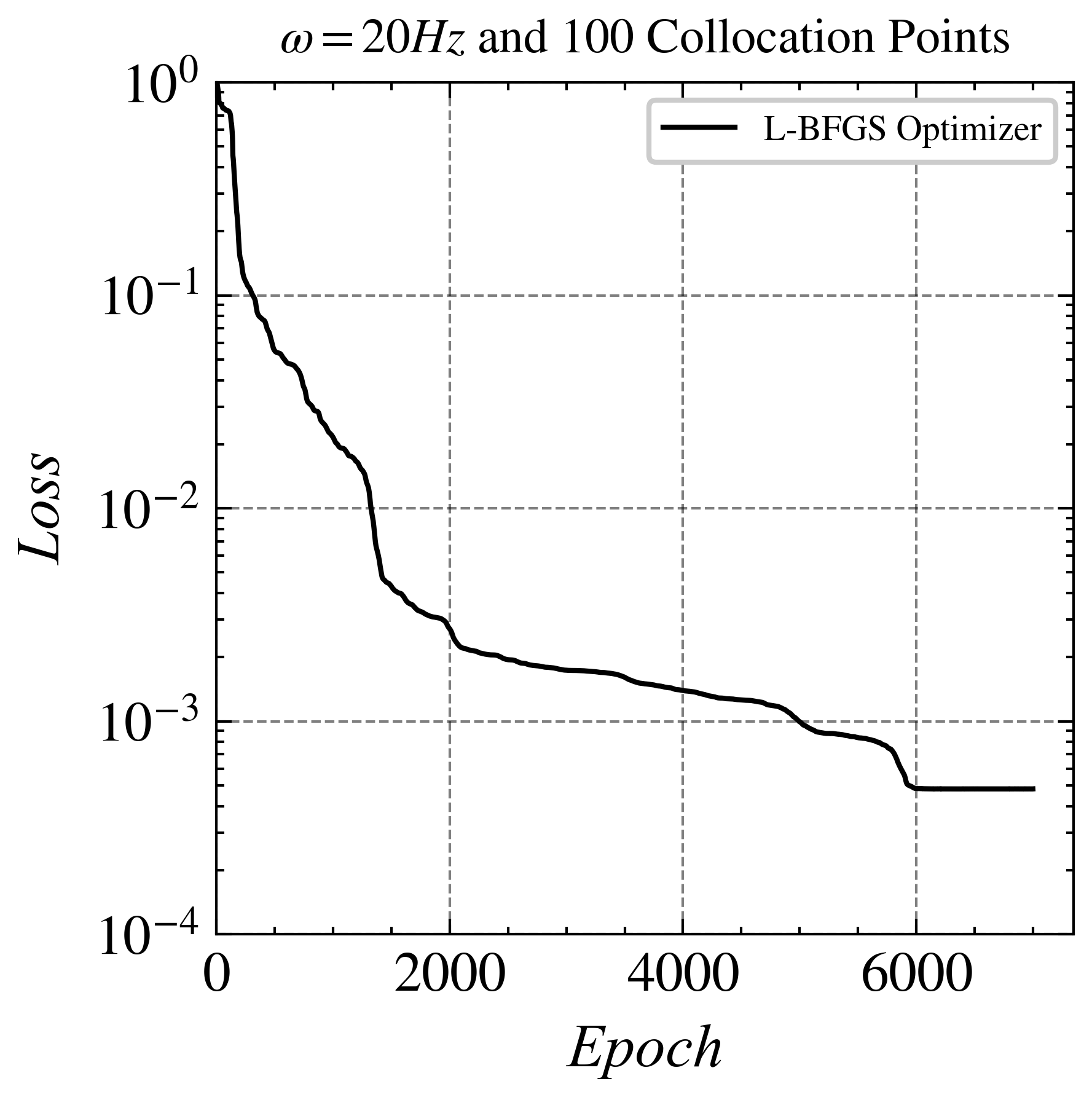}
    \caption{Loss curve with L-BFGS Optimizer over epochs}
    \label{fig:20hz_pinn_b}
  \end{subfigure}

  \vspace{10pt} % Add vertical space between the rows

  \begin{subfigure}{0.45\textwidth}
    \centering
    \includegraphics[width=\linewidth]{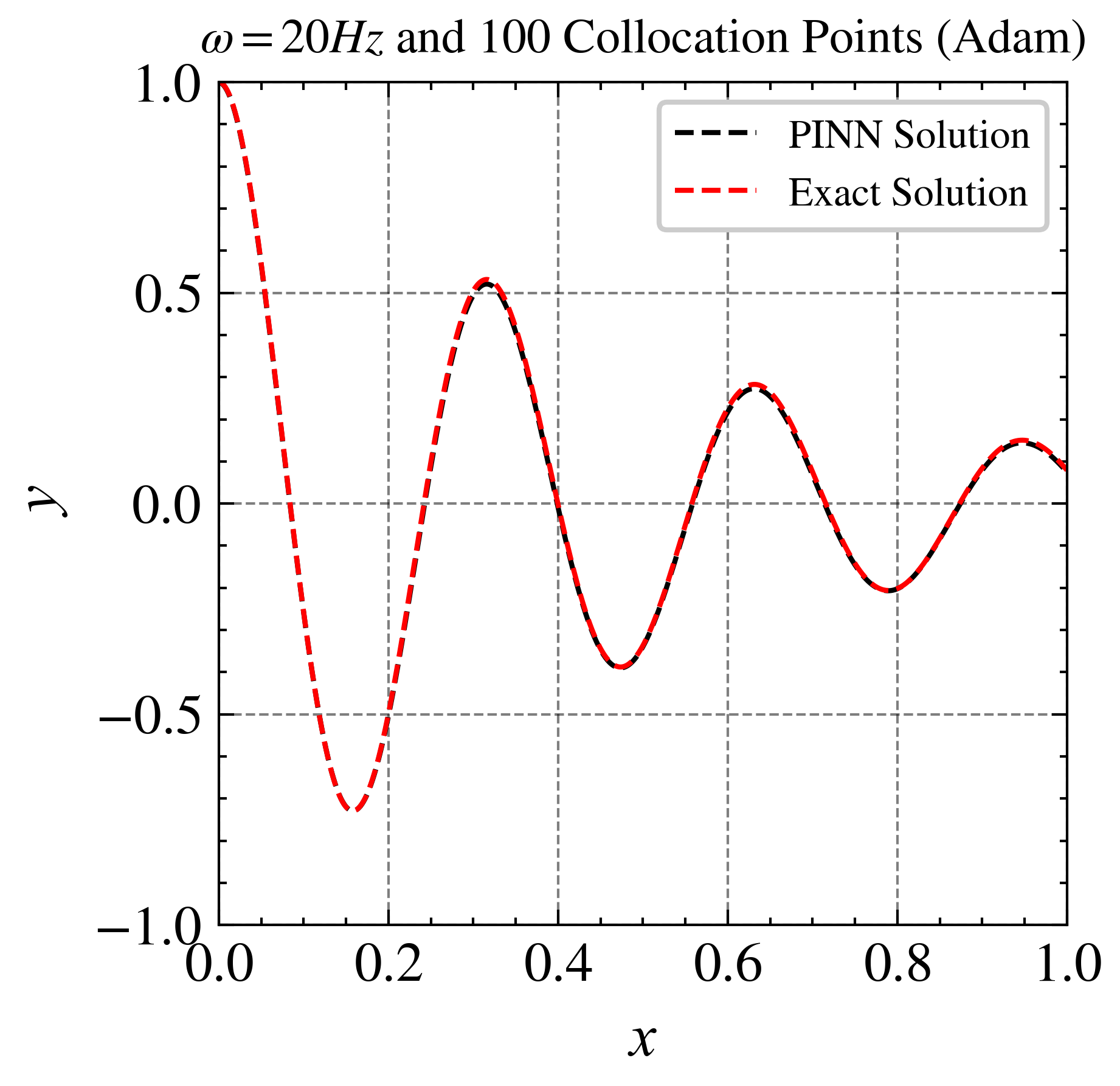}
    \caption{PINN solution vs. exact solution using Adam}
    \label{fig:20hz_pinn_c}
  \end{subfigure}
  \hfill
  \begin{subfigure}{0.45\textwidth}
    \centering
    \includegraphics[width=\linewidth]{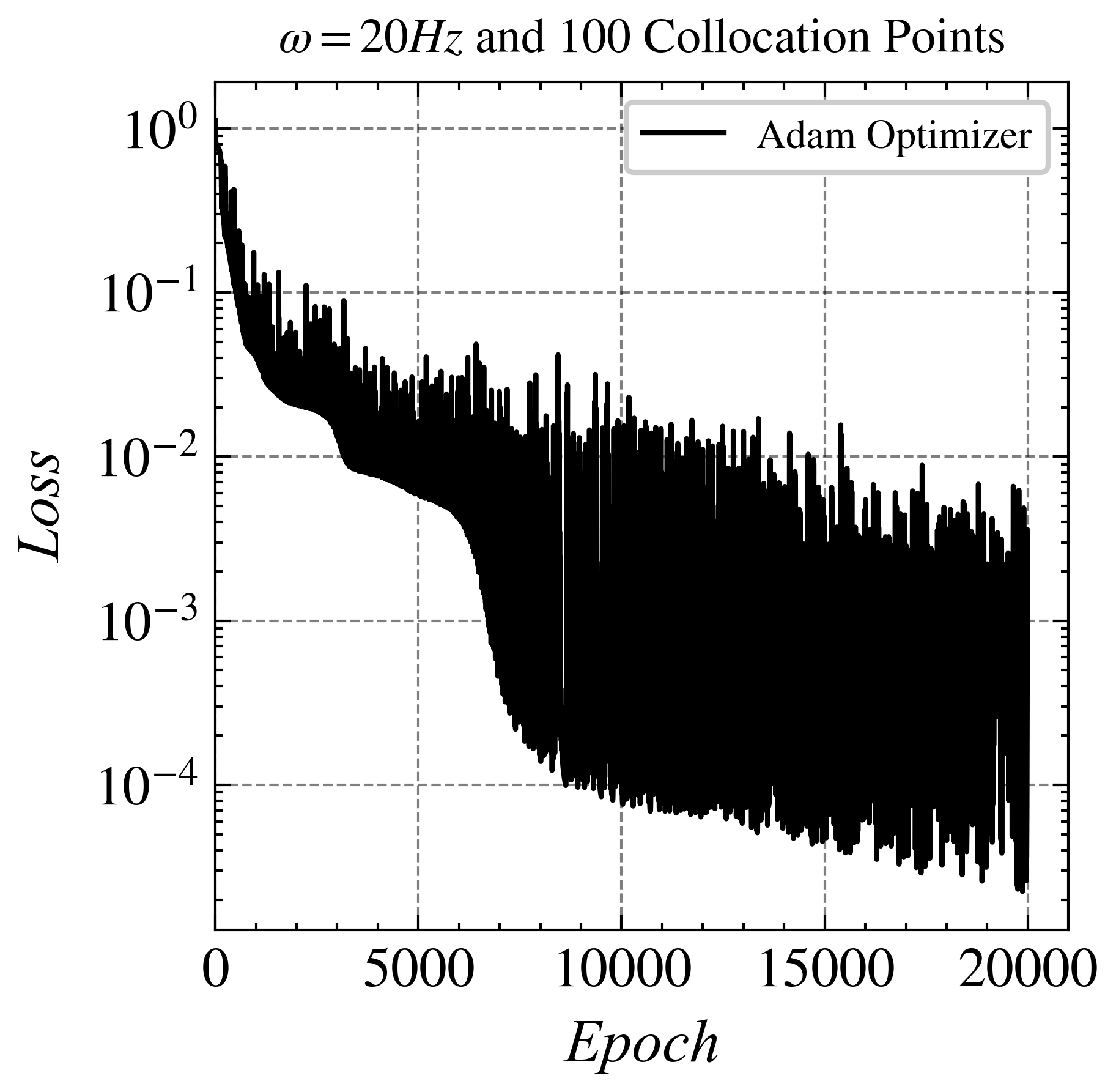}
    \caption{Loss curve with Adam Optimizer over epochs}
    \label{fig:20hz_pinn_d}
  \end{subfigure}

  \caption{Comparisons of Adam optimizer vs. L-BFGS optimizer at 20Hz}
  \label{fig:20hz_pinn}
\end{figure}

%\subsubsection{W0 as 20hz}
For $\omega_0 = 20$ [Figure. \ref{fig:20hz_pinn_a}], PINN was able to fit well where the loss reaches the order of $10^{-3}$ with both Adam and L-BFGS optimizers. With the LBFGS optimizer, it converged at around 2200 epochs, while with the Adam optimizer, it converged at around 7000 epochs. 

%\subsubsection{W0 as 30hz}
For $\omega_0 = 30$, PINN was also able to reach the order of $10^-3$ loss with both Adam and L-BFGS optimizers. With the L-BFGS optimizer, it converged at around 7000 epochs. The Adam optimizer needs around 22,000 epochs to converge.

\begin{figure}[H]
  \centering
  \begin{subfigure}{0.45\textwidth}
    \centering
    \includegraphics[width=\linewidth]{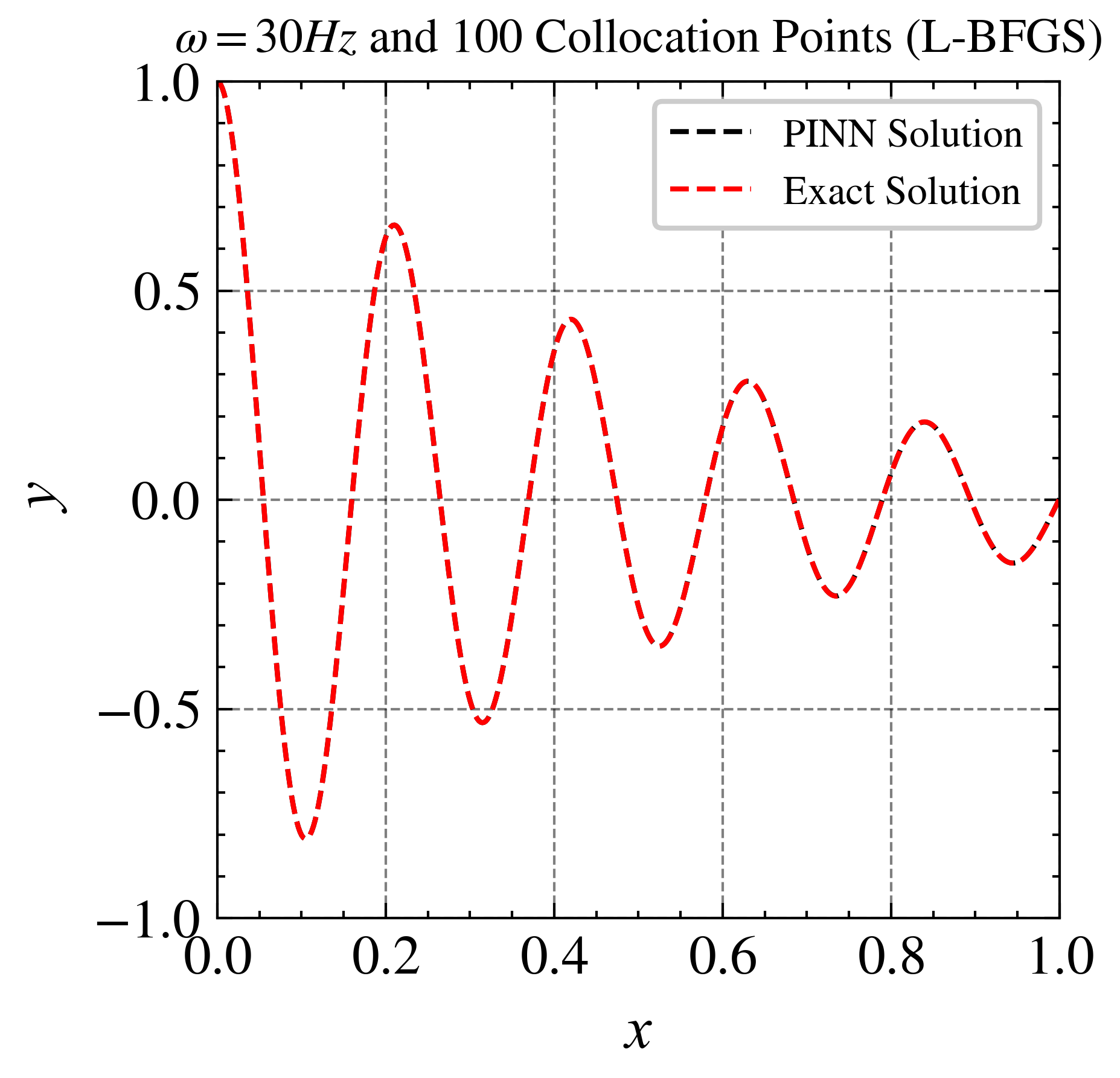}
    \caption{PINN solution vs. exact solution using L-BFGS}
    \label{fig:30hz_pinn_a}
  \end{subfigure}
  \hfill
  \begin{subfigure}{0.45\textwidth}
    \centering
    \includegraphics[width=\linewidth]{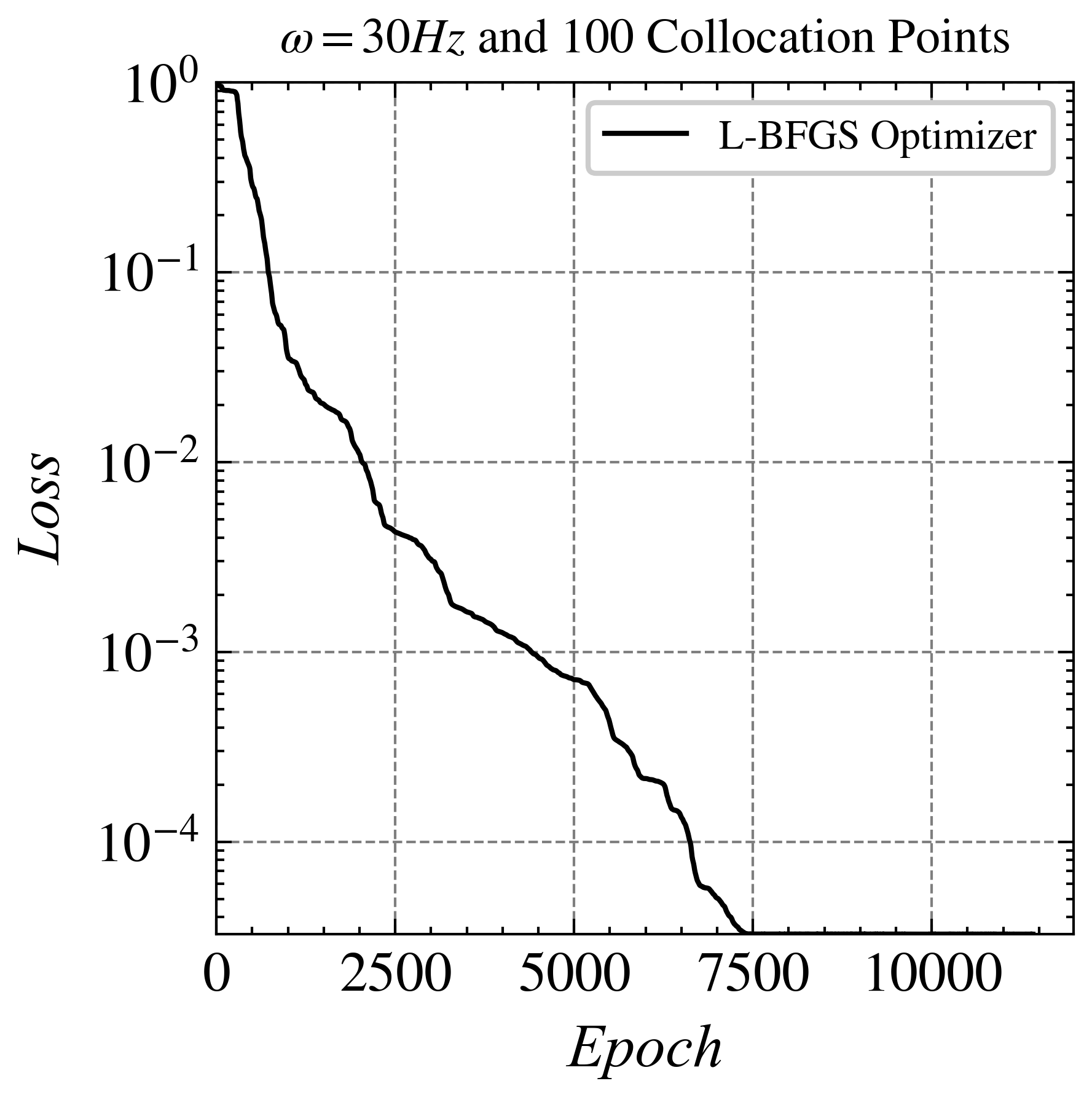}
    \caption{Loss curve with L-BFGS Optimizer over epochs}
    \label{fig:30hz_pinn_b}
  \end{subfigure}
  
  \vspace{10pt} % Add vertical space between the rows

  \begin{subfigure}{0.45\textwidth}
    \centering
    \includegraphics[width=\linewidth]{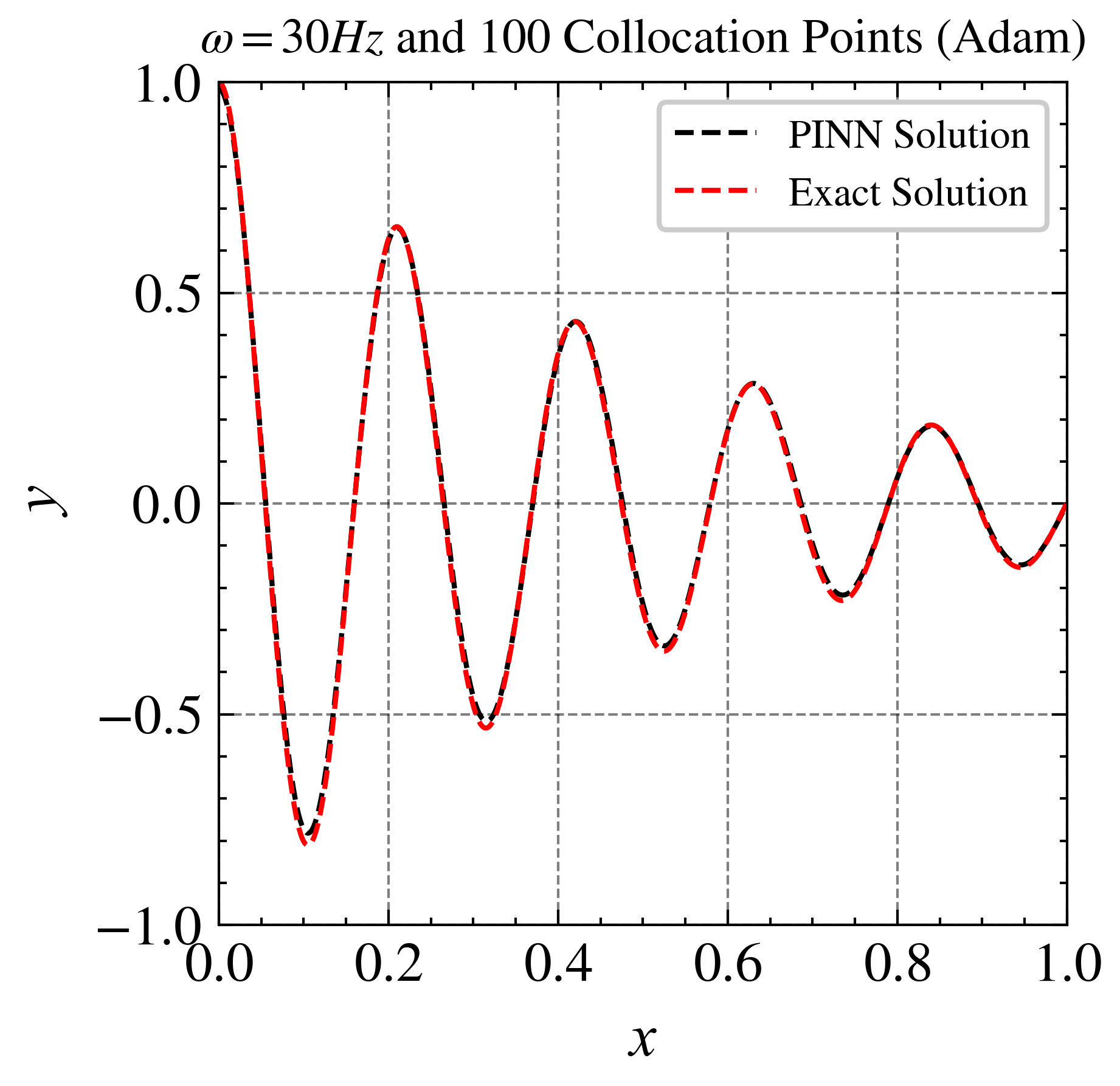}
    \caption{PINN solution vs. exact solution using Adam}
    \label{fig:30hz_pinn_c}
  \end{subfigure}
  \hfill
  \begin{subfigure}{0.45\textwidth}
    \centering
    \includegraphics[width=\linewidth]{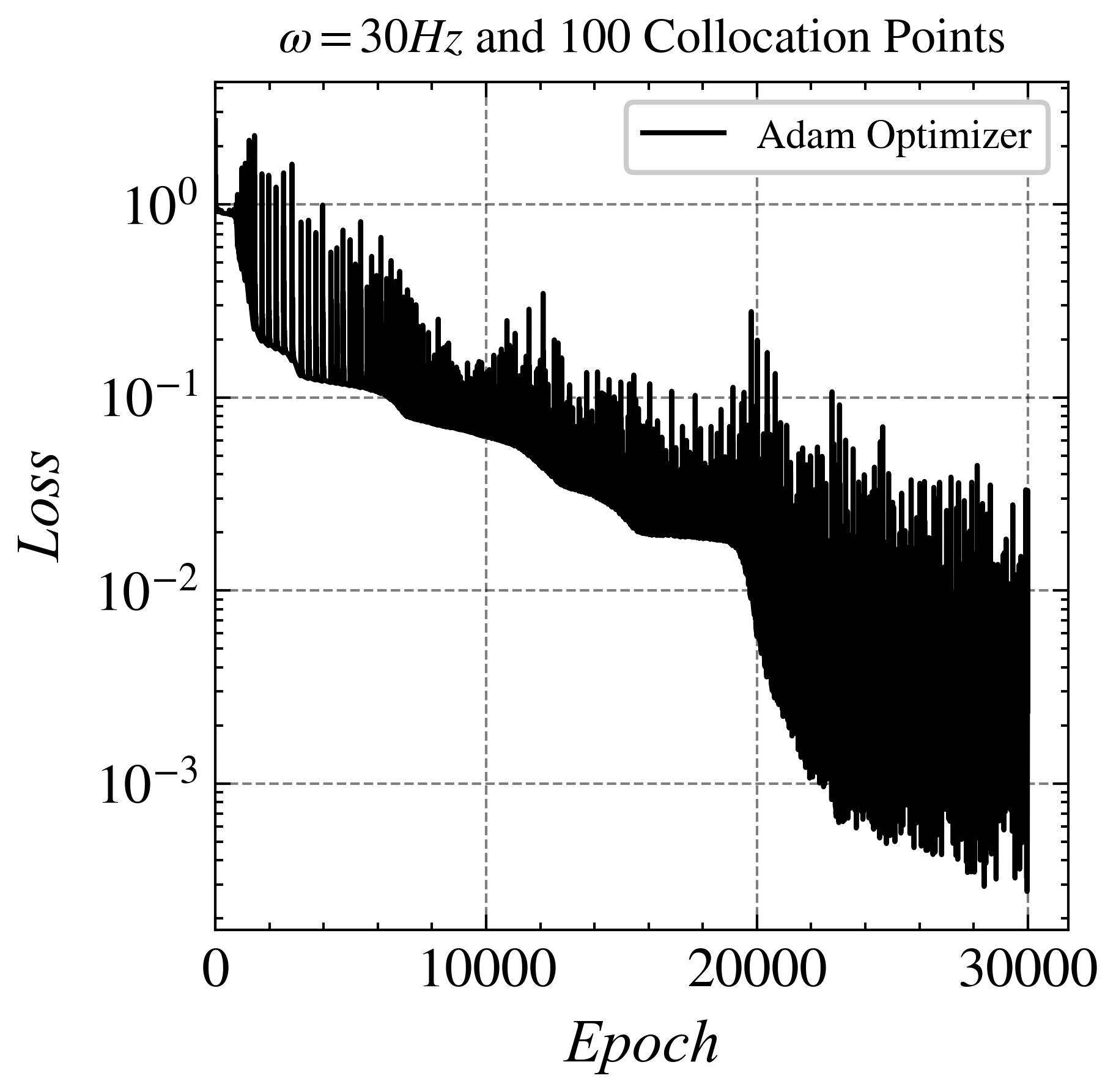}
    \caption{Adam Loss}
    \label{fig:30hz_pinn_d}
  \end{subfigure}

  \caption{Comparison of the Adam optimizer with the L-BFGS optimizer at 30Hz: (a) comparing the PINN solution and the exact solution using L-BFGS, (b) visualization of loss curve with L-BFGS Optimizer across epochs, (c) analyzing the PINN solution and exact solution using Adam, (d)  Loss using Adam Optimizer.}
\end{figure}

%\subsubsection{W0 as 40hz}
In comparing the convergence behaviour of the Adam optimizer and the L-BFGS optimizer at a frequency of 40 Hz, it becomes apparent that both algorithms exhibit different characteristics and performances, particularly in their speed of convergence and stability during optimization. The Adam optimizer eventually converges; however, this is achieved after a significant number of iterations. This observation raises concerns as we transition to higher frequencies, suggesting challenges in PINN convergence. This could be partly attributed to the well-documented issue of spectral bias inherent in neural networks.

When considering which optimizer to use, it's crucial to select one with caution as it can have a significant impact on the efficiency of the training process. We found out that using Adam with L-BFGS gave the best results.

\begin{figure}[H]
  \centering
  \begin{subfigure}{0.45\textwidth}
    \centering
    \includegraphics[width=\linewidth]{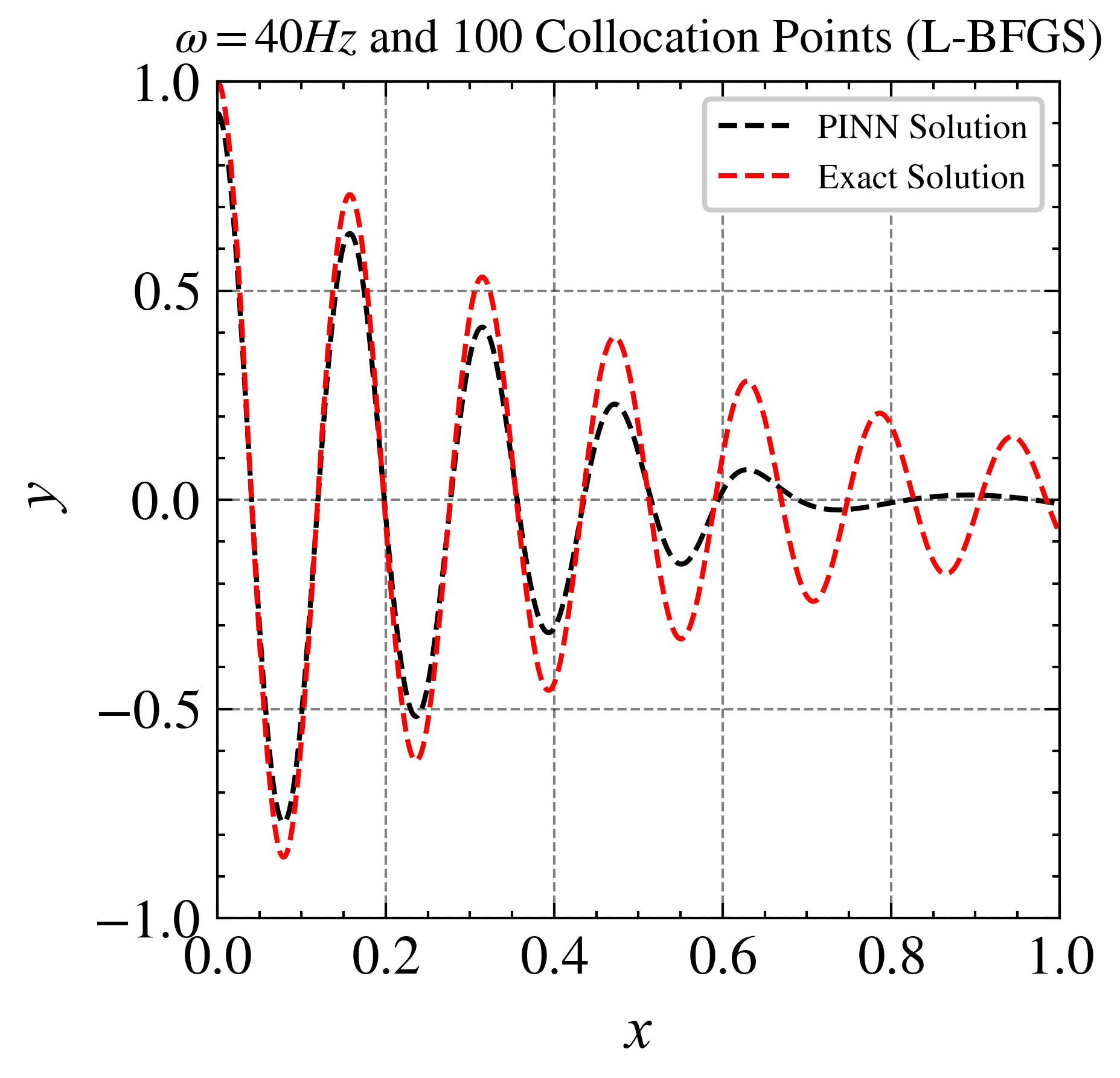}
    \caption{PINN solution vs. exact solution using L-BFGS}
    \label{fig:40hz_pinn_a}
  \end{subfigure}
  \hfill
  \begin{subfigure}{0.45\textwidth}
    \centering
    \includegraphics[width=\linewidth]{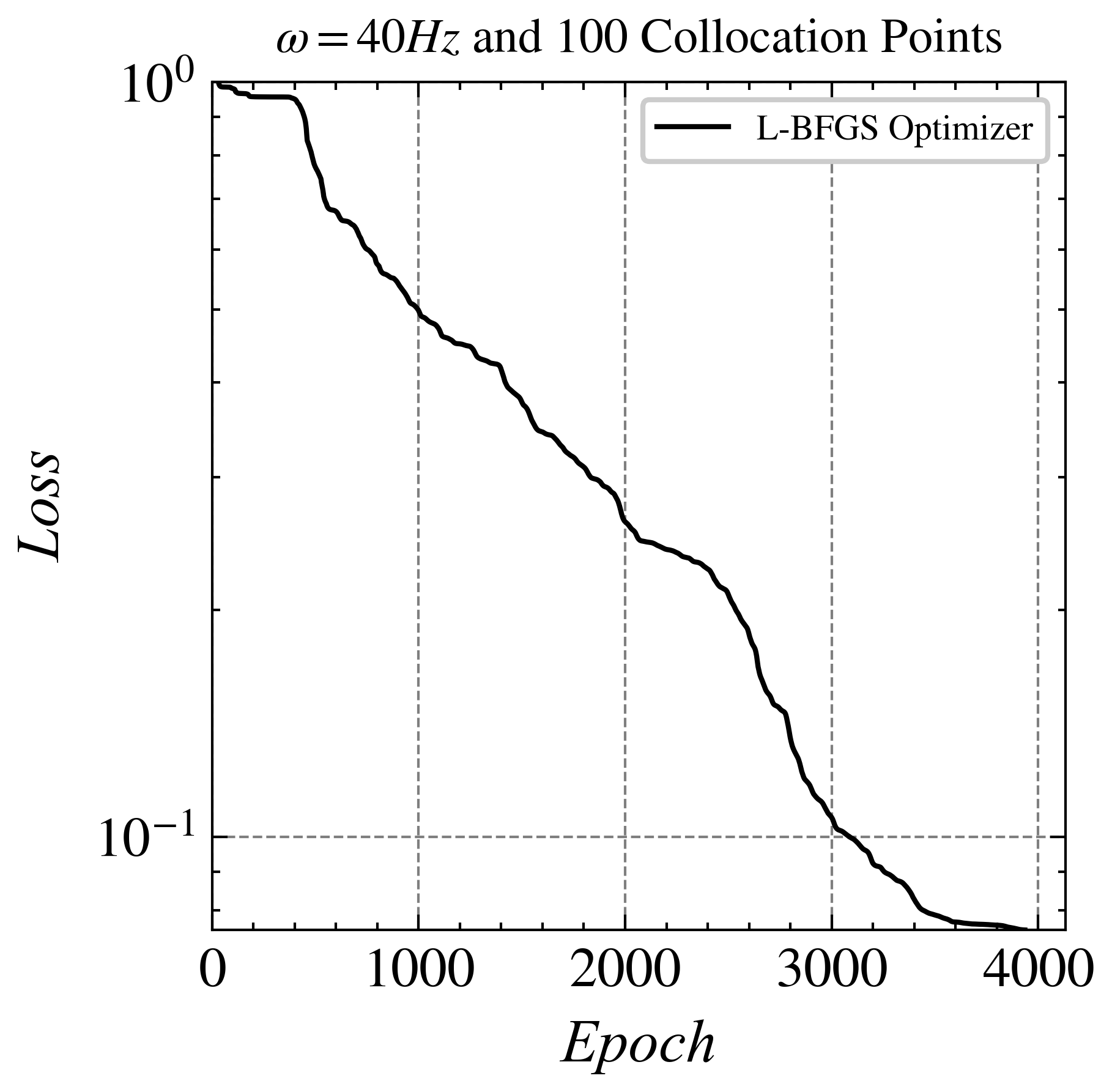}
    \caption{Loss curve with L-BFGS Optimizer over epochs}
    \label{fig:40hz_pinn_b}
  \end{subfigure}

  \vspace{10pt} % Add vertical space between the rows

  \begin{subfigure}{0.45\textwidth}
    \centering
    \includegraphics[width=\linewidth]{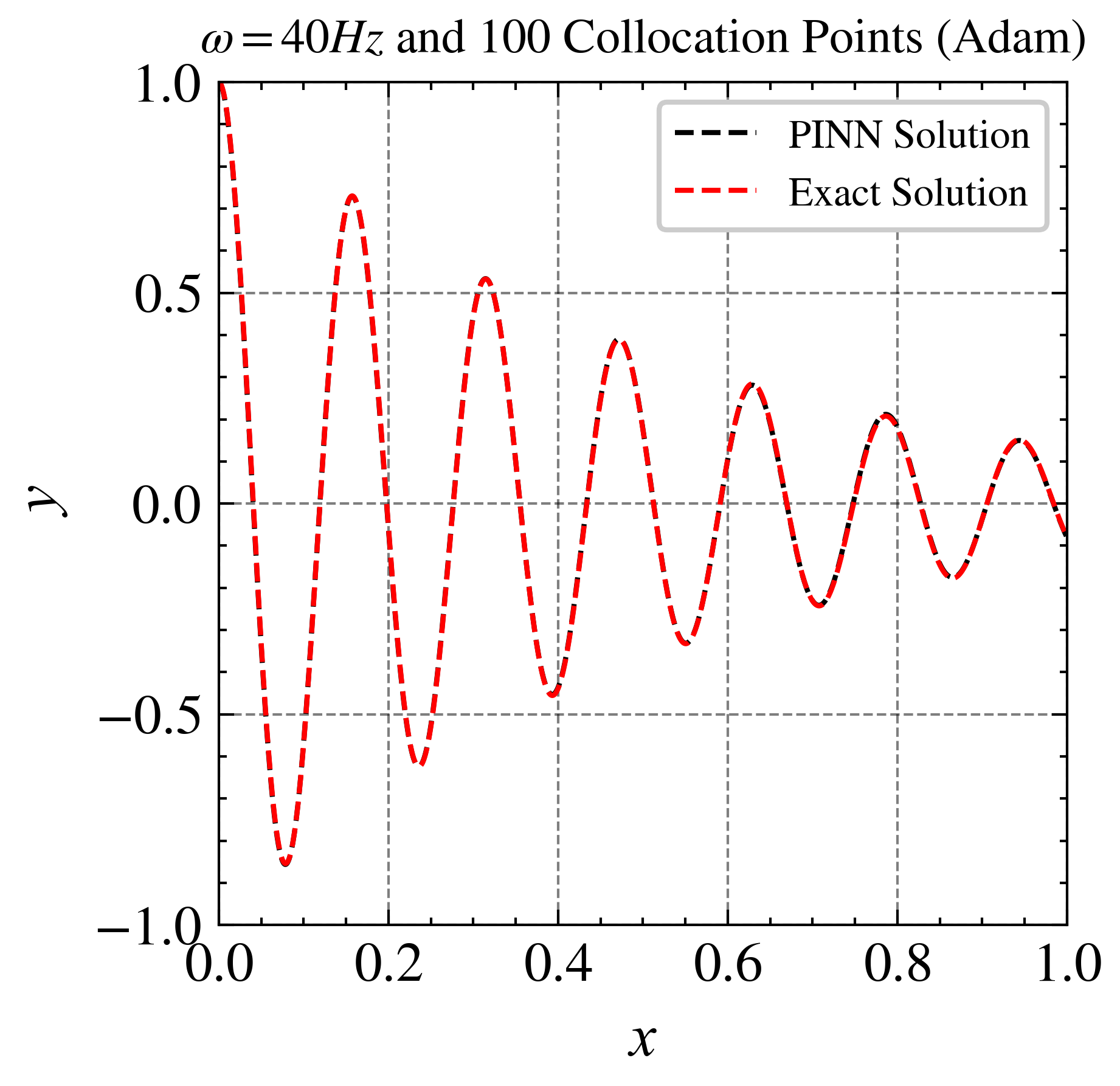}
    \caption{PINN solution vs. exact solution using Adam}
    \label{fig:40hz_pinn_c}
  \end{subfigure}
  \hfill
  \begin{subfigure}{0.45\textwidth}
    \centering
    \includegraphics[width=\linewidth]{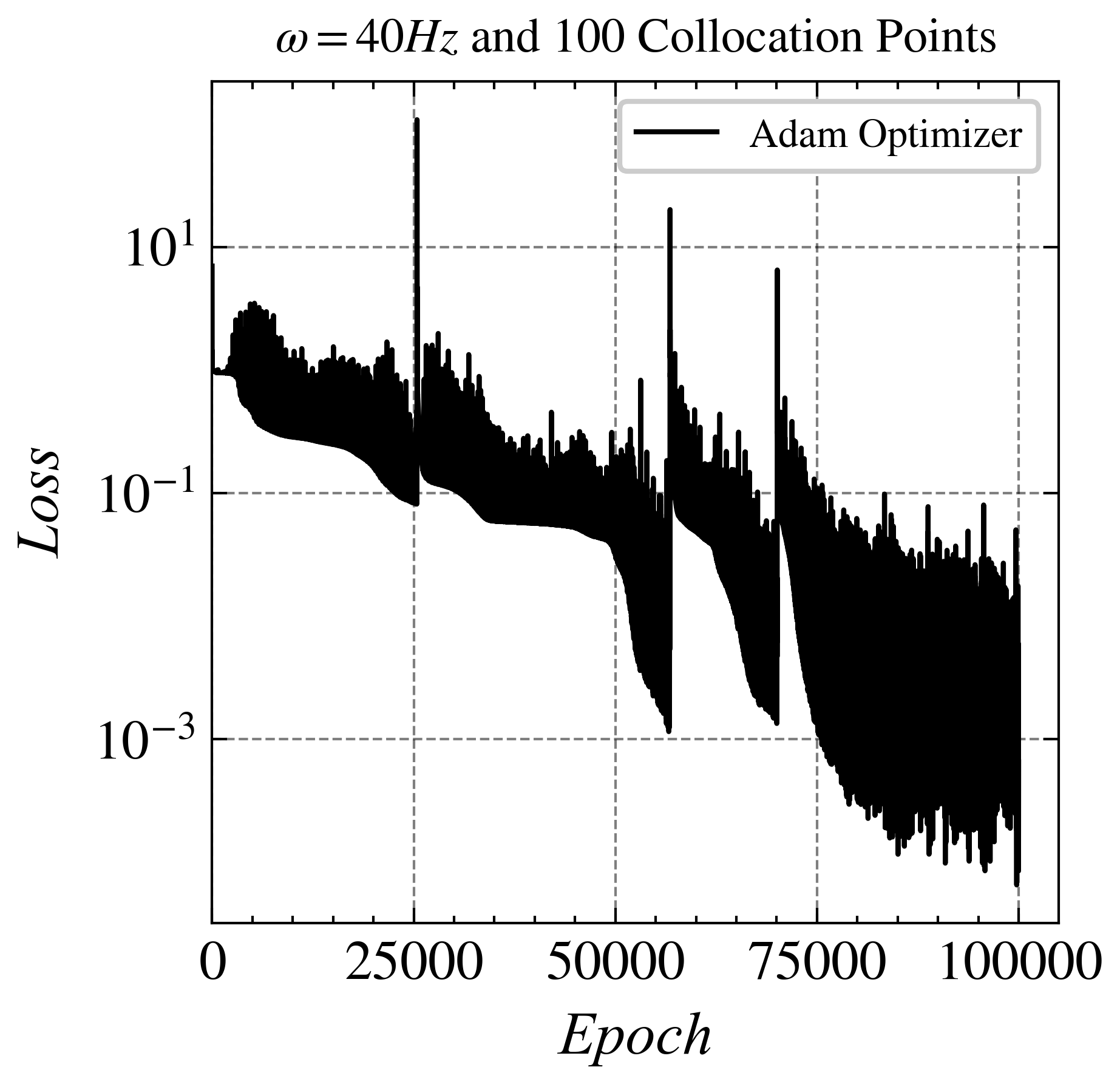}
    \caption{Loss curve with Adam Optimizer over epochs}
    \label{fig:40hz_pinn_d}
  \end{subfigure}

  \caption{Comparisons of Adam optimizer vs. L-BFGS optimizer at 40Hz}
  \label{fig:40hz_pinn}
\end{figure}

%\subsection{Discussion of the Results Obtained from the Two Optimizers}
The performance of PINN is observed to be consistent in two different frequency scenarios (20 Hz and 30 Hz) when using  Adam and L-BFGS. This indicates that the quality of predictions remains stable regardless of the chosen optimization algorithm. It's important to note that both optimizers ultimately achieve convergence and deliver favorable predictive outcomes; however, they exhibit notable differences in behavior.

The Adam optimizer, although effective, requires a higher number of iterations to reach convergence and shows some degree of instability compared to the L-BFGS optimizer. At times, Adam outperforms L-BFGS, possibly due to L-BFGS temporarily getting stuck in a local minimum leading to quick convergence. In the context of the mentioned frequency scenarios, a learning rate of 0.1 was set for the L-BFGS optimizer. Since L-BFGS is a quasi-newton method, it depends on the initial guess. 

At a frequency of 40 Hz,  Adam optimizer was able to solve the problem but it required significantly more number of iterations, with almost 80,000 iterations needed to reach convergence. On the other hand, LBFGS failed to converge and seemed to fit the lower-frequency components of the problem. It is important to note that the loss in this case remained at the order of $10^{-1}$, highlighting the problems of solving high-frequency cases within the PINN framework.

\subsubsection{Transfer Learning}
This section introduces a transfer learning technique to boost the robustness and convergence of training PINN. Transfer learning presents a promising solution to mitigate these issues by leveraging the pre-trained model or the baseline PINN, thereby furnishing an advantageous initial guess to expedite convergence. To assess the efficacy of this technique, we conducted a series of experiments involving different optimization algorithms and compared their performances respectively.  The baseline low-frequency model is required to initiate the transfer learning of PINN from low-frequency to high-frequency. The models mentioned in the previous part are selected for transfer learning to facilitate the scaling of the model to higher frequencies. The baseline model is established, revealing that as the frequency is elevated, the capability of the PINN, given the present configuration, to scale effectively diminishes.  It is important to note that LBFGS, a Newton-based optimization method, exhibits sensitivity to the initial guess, rendering it susceptible to convergence challenges, including the risk of getting trapped in local minima or failing to converge even after a substantial number of iterations.  Some empirical evidence shows that Adam optimizer when used with a combination of the L-BFGS optimizer, ensures that the latter escapes from the local minima \cite{markidis2021old}. We selected the baseline models at 30Hz generated by both the Adam and LBFGS optimizers. These models were subsequently employed as the starting point for training a PINN model targeting a frequency of 40 Hz. This approach enables us to evaluate which of the two optimizers produces a more effective baseline model for this task.

\subsubsection{Discussion on results}
In this section, we test out both Adam and L-BFGS optimizers, to see which of the two performs better as a source model to scale to higher frequencies. \\

\begin{figure}[H]
\centering

\begin{subfigure}{0.45\textwidth}
  \centering
  \includegraphics[width=1\linewidth]{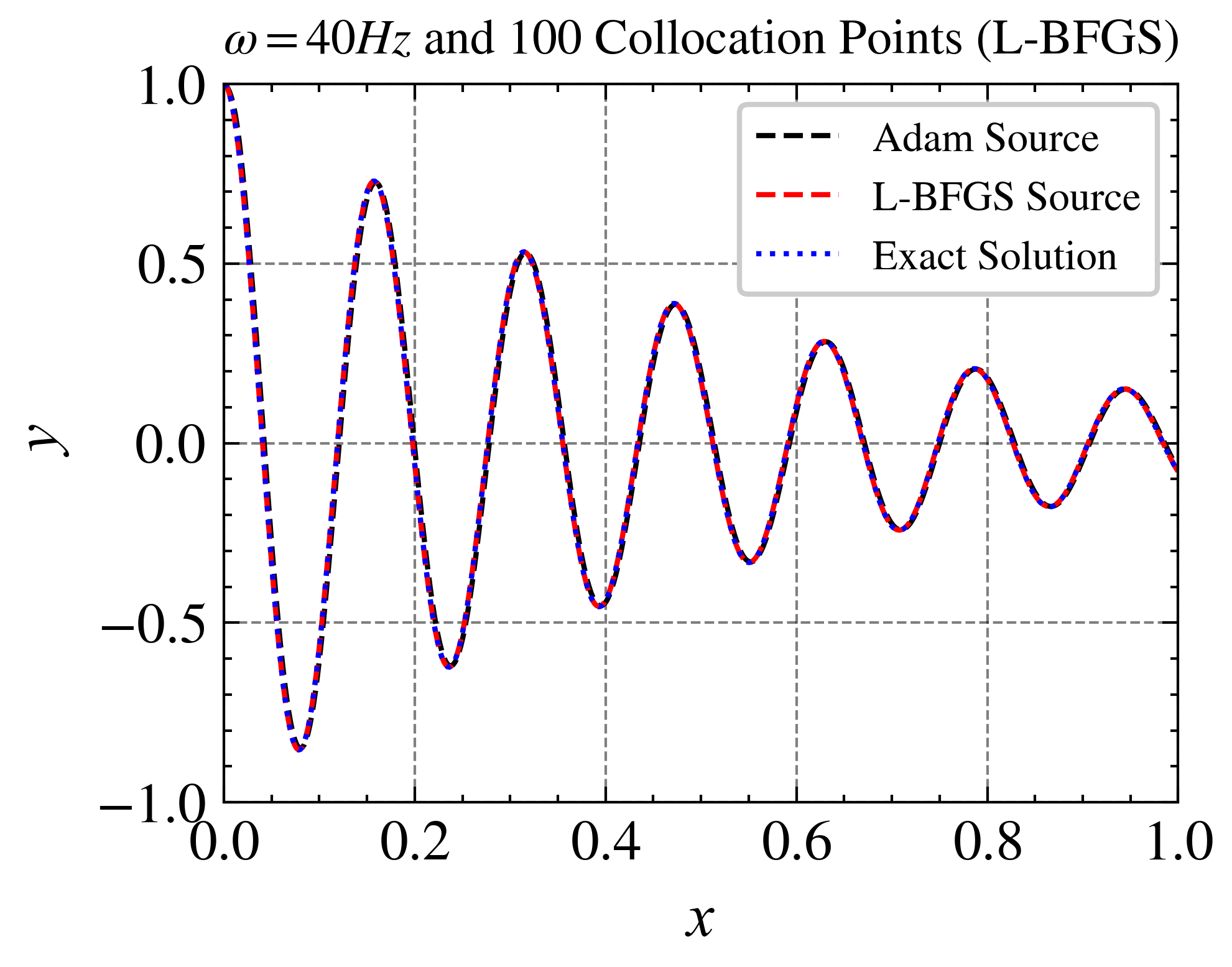}
  \caption{PINN vs Exact Solution}
  \label{fig2}
\end{subfigure}
\begin{subfigure}{0.45\textwidth}
  \centering
  \includegraphics[width=1\linewidth]{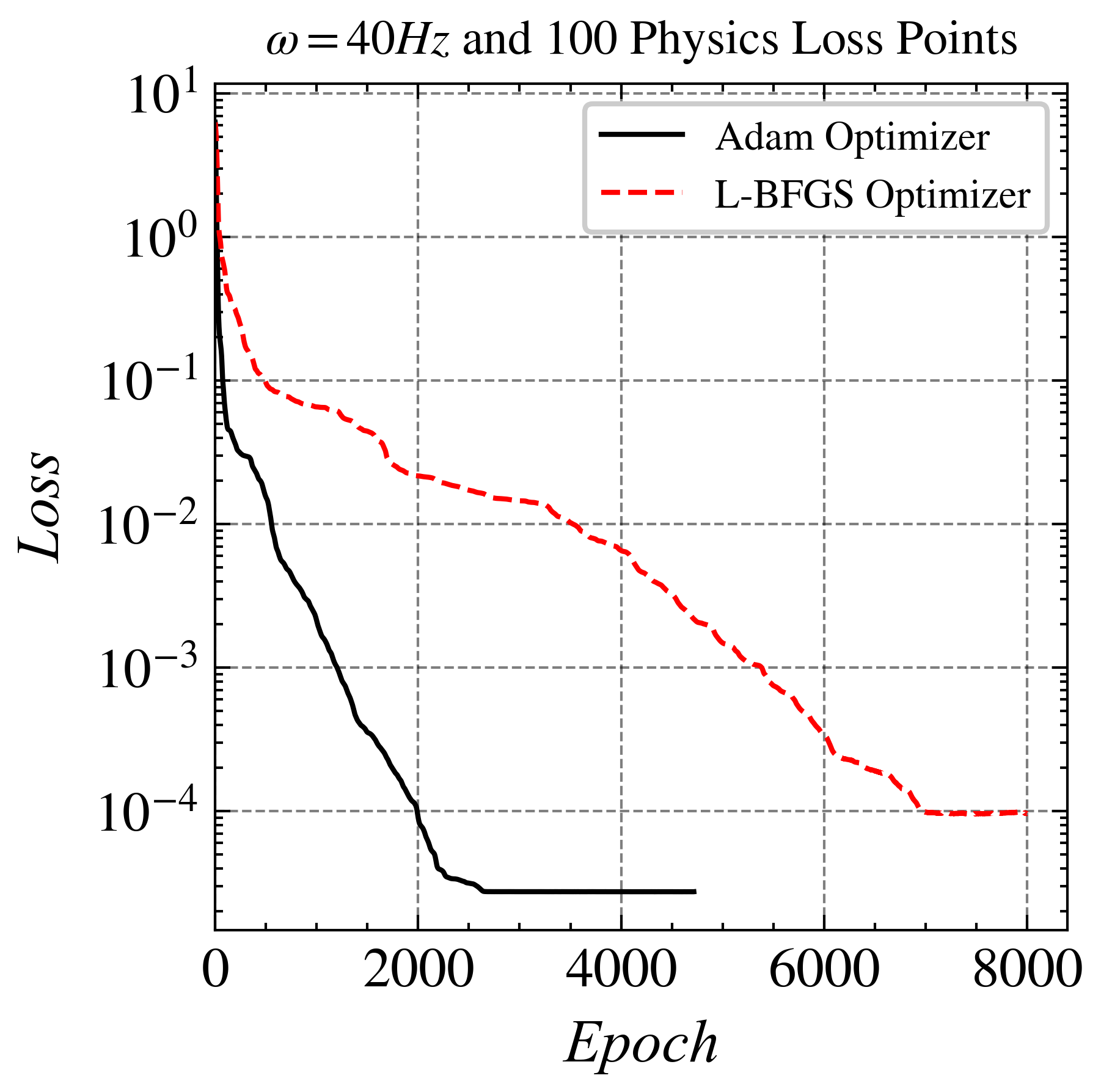}
  \caption{Adam vs L-BFGS Initializers.}
  \label{40hz_loss_compare}
\end{subfigure}
\caption{40Hz comparisons for Adam and LBFGS. }
\end{figure}

In the following results, we use L-BFGS to train the network. We make use of transfer learning and compare Adam and L-BFGS baseline models. In Fig[\ref{40hz_loss_compare}], it is evident that the source model for 30 Hz with Adam performed much better than that of L-BFGS. As mentioned above, this might be due to the nature of L-BFGS. \\

When we compare the results of the 40Hz case without transfer learning (Fig. \ref{fig:40hz_pinn_d}) with those using transfer learning (Fig. \ref{40hz_loss_compare}), we observe that Adam achieved a loss order of $10^{-3}$ in about 75000 iterations, while the one using transfer learning achieved the same order in less than 2000 epochs. This not only reduced the computation time significantly but also provided a more accurate solution.

Now, moving on to the next set of results, we use the 40Hz model as the source to train the model on $\omega=50$. Similarly, we used the $\omega=50$ model as the source to train the model on $\omega=60$.

\begin{figure}[H]
  \begin{subfigure}{0.45\textwidth}
    \centering
    \includegraphics[width=1\linewidth]{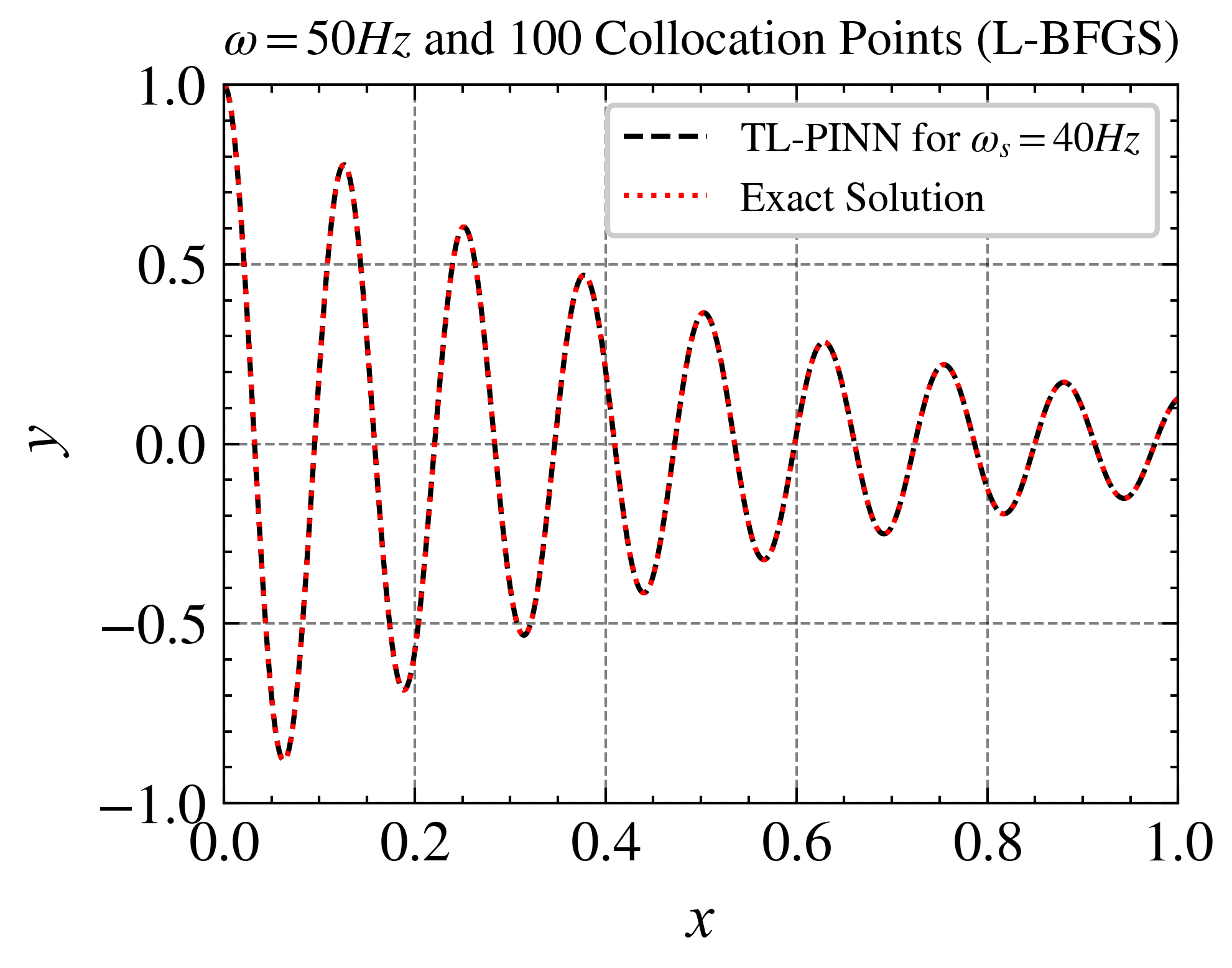}
    \caption{Source PINN: 40 Hz}
    \label{50hz_solution}
  \end{subfigure}
  \hfill
  \begin{subfigure}{0.45\textwidth}
    \centering
    \includegraphics[width=1\linewidth]{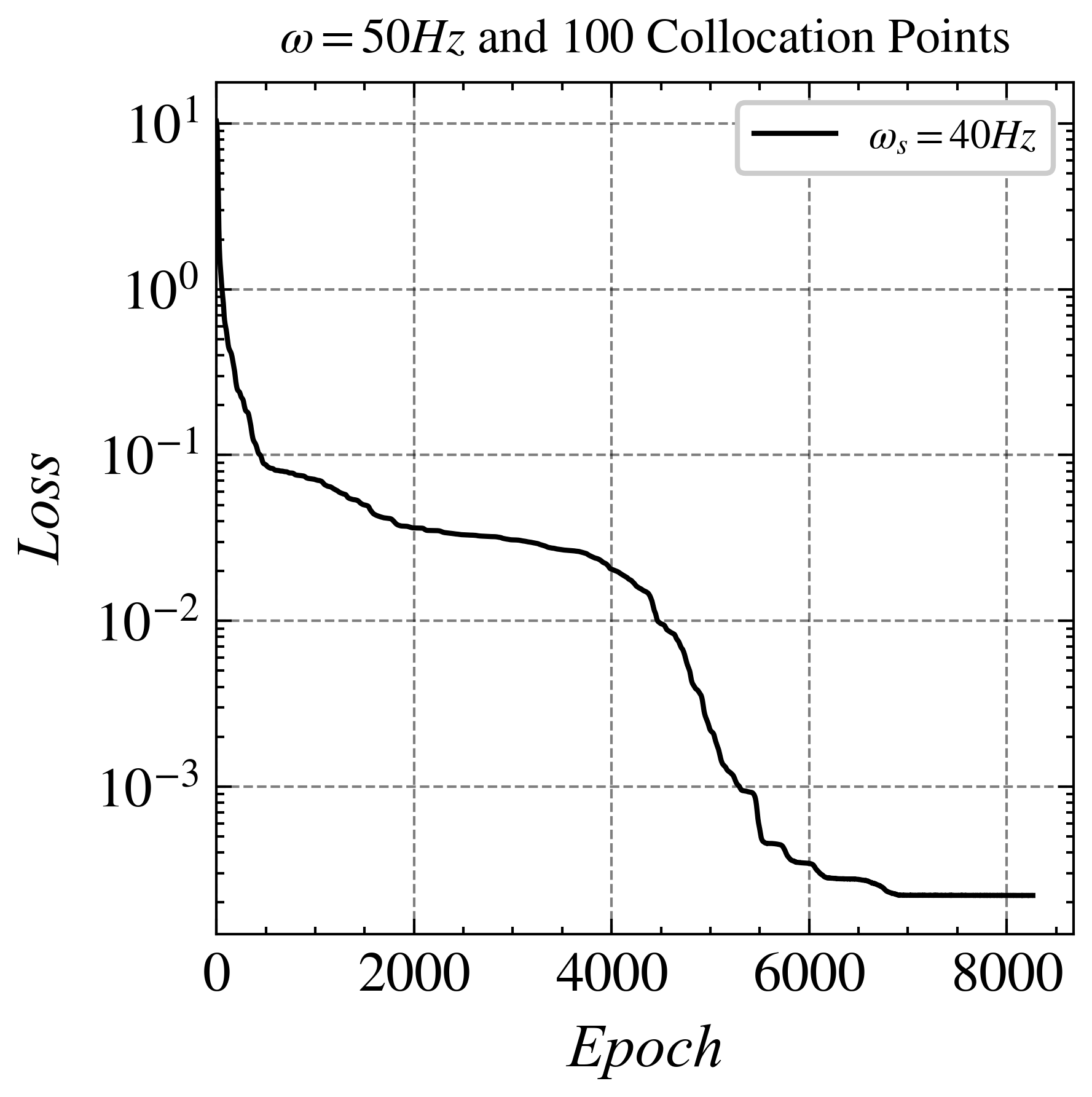}
    \caption{Loss}
    \label{50hz_loss}
  \end{subfigure}

  \vspace{0.5cm}  % Adjust the vertical space between the first and second rows of subfigures

  \begin{subfigure}{0.45\textwidth}
    \centering
    \includegraphics[width=1\linewidth]{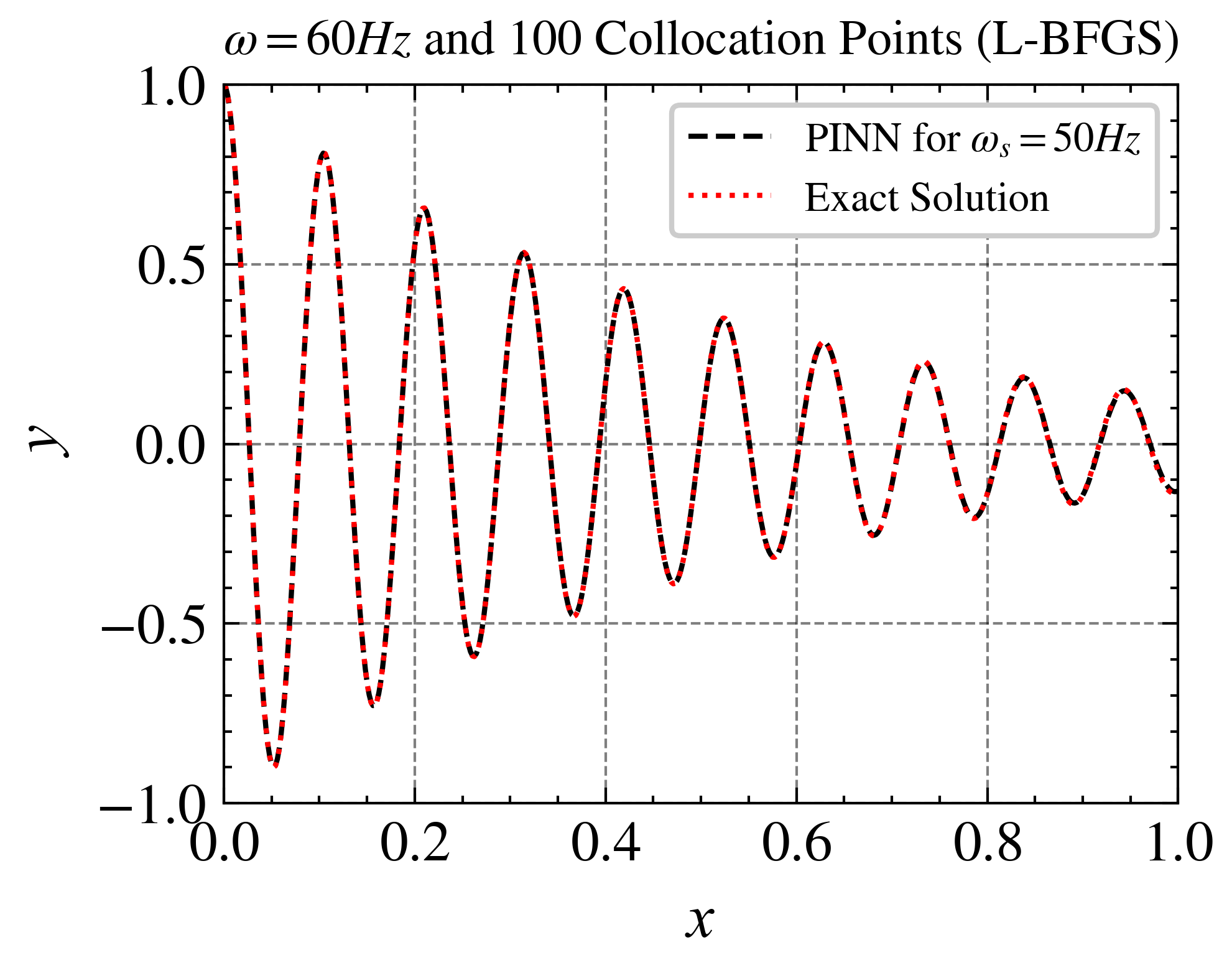}
    \caption{Source PINN: 50 Hz}
    \label{60hz_solution}
  \end{subfigure}
  \hfill
  \begin{subfigure}{0.45\textwidth}
    \centering
    \includegraphics[width=1\linewidth]{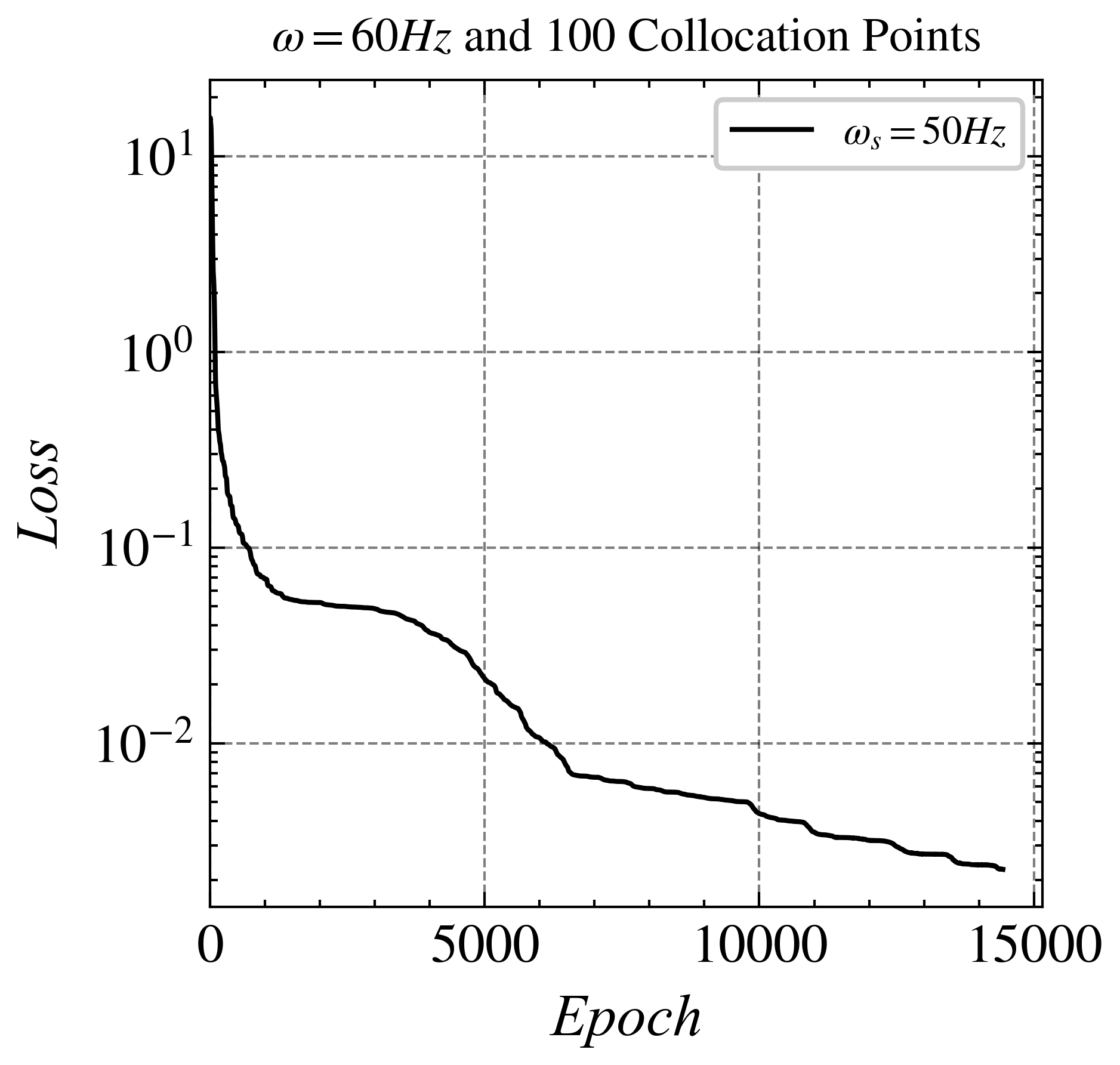}
    \caption{Loss}
    \label{60hz_loss}
  \end{subfigure}

  \centering
  \vspace{0.5cm}  % Adjust the vertical space between the second and third rows of subfigures
  \caption{Transfer Learning Results for $\omega = 50Hz$ and $\omega = 60Hz$. In both the cases, L-BFGS optimizer was used to train the model. The order of loss for $\omega = 60Hz$ is higher than that that of $\omega = 50Hz$ because of the complexity of the solution, however the PINN solutions fit well with the exact solution.}
\end{figure}

\begin{table}[htbp]
    \centering
    \fbox{
    \begin{tabular}{lcccc}
        \multicolumn{4}{c}{\textbf{L2 Relative Error (\%) of Adam, LBFGS, and Transfer Learning}} \\
        \toprule
         & \textbf{Adam} & \textbf{LBFGS} & \textbf{Transfer Learning (Adam)} \\
        \midrule
        \textbf{20} & 0.95 & 0.92  & - \\
        \textbf{30} & 1.24 & 0.88  & - \\
        \textbf{40} & 0.80 & 0.95  & 0.037  \\
        \textbf{50} & 3.15 & 0.95 & 0.881   \\
        \textbf{60} & 71.54 & 0.95 & 0.218 \\
        \bottomrule
    \end{tabular}
    }
    \caption{Relative L2 Error (\%).}
    \label{tab:results}
\end{table}

\subsection{1D Wave Equation}

The wave equation:
\[
\frac{\partial^2 u}{\partial t^2} = c^2 \nabla^2 u
\]
The equation models the oscillations of a one-dimensional string (\(u = u(x, t)\)), the oscillations of a two-dimensional thin membrane (\(u = u(x, y, t)\)), or the pressure oscillations of an acoustic wave in air (\(u = u(x, y, z, t)\)). The constant \(c\) denotes the velocity of wave propagation for the oscillations and is also known as the wave velocity in certain literature.

Although typically discussed in just one spatial dimension (\(x\)) due to time (\(t\)) being the only independent variable, it's important to mention that the variable we're studying (\(u\)) can represent movement in another direction, like up and down (\(y\)). For example, this occurs when a string is not only moving horizontally (\(x\)) but also vertically (\(y\)), as seen on a flat surface.

The unknown function \(u\) depends on space \(x\) and time \(t\), and can be represented as an equation: 
\begin{equation}
u = u(x, t)
\end{equation}

To solve the function, we need also Initial conditions and boundary conditions. In the experiments, we use the following conditions. 

\begin{align}
\begin{cases}
    u_{tt} = c^2  u_{xx} \\
    u(x, 0) = \sin(x) \\
    u_t(x, 0) = \sin(x) \\
    u_b(0, t) = u_b(\pi, t) = 0
\end{cases}
& \text{for } 0 \leq t \leq 2\pi, \quad 0 \leq x \leq \pi \
\end{align}

We solve the case where \(c=1\). Specifically, we address the equation with homogeneous Dirichlet conditions, \(c=1\), and compare the results with the analytical solution.

As we increase \(c\) from 1 to 2, we observe that the solution takes much longer to converge. To address this, we employ transfer learning. We first train the model for \(c=1\) and then use this knowledge to approximate the solution for \(c=2\).

To do so, we approximate the underlying solution with a feedforward dense neural network with tunable parameters \(\theta\):
\[
\hat{u}_\theta(x, t) \approx u(x, t)
\]

This approach allows us to efficiently model and compare solutions under different conditions, providing a deeper understanding of the system's behaviour.

The loss function is given by
\[
L(\theta) = W_F L_F(\theta) + W_I L_I(\theta) + W_B L_B(\theta),
\]
with $\theta$ as the weights of neural networks.

The interior residual is given by
\begin{equation}
r_{\text{int},\theta}(x,t):= \hat{u}_{\theta, tt}(x,t) - c^2(\hat{u}_{\theta, xx}(x,t)), \quad \forall ~t \in [0,T],~ x \in [0,\pi].
\end{equation}

The spatial boundary residual or boundary conditions are given by
\begin{equation}
\begin{aligned}
    &r_{\text{sb},\theta}(0,t):= \hat{u}_{\theta}(0,t)-  u_b(0,t), \\
    &r_{\text{sb},\theta}(\pi,t):= \hat{u}_{\theta}(\pi,t)- u_b(\pi,t), \\
    &\forall t \in (0, T].
\end{aligned}
\end{equation}

The temporal boundary residual is given by
\begin{equation}
\begin{aligned}
    &r_{\text{tb},\theta}(x):= \hat{u}_{\theta}(x,0) - u(x,0), \quad \forall x \in [0,\pi], \\
    &r_{\text{tb},\theta}(x):= \hat{u}_{\theta,t}(x,0) - u_t(x,0), \quad \forall x \in [0,\pi].
\end{aligned}
\end{equation}

With the training input points corresponding to low-discrepancy Sobol sequences, the loss terms are:

\begin{align}
L_{F}(\theta) &= \frac{1}{N_{\text{int}}}\sum_{i=1}^{N_{\text{int}}} r_{\text{int},\theta}^2(x_i,t_i) 
\\
L_{B}(\theta) &= \frac{1}{N_{\text{sb}}}\sum_{i=1}^{N_{\text{sb}}} r_{\text{sb},\theta}^2(t_i,0) + \frac{1}{N_{\text{sb}}}\sum_{i=1}^{N_{\text{sb}}} r_{\text{sb},\theta}^2(t_i,\pi), 
\\
L_{I}(\theta) &= \frac{1}{N_{\text{tb}}}\sum_{i=1}^{N_{\text{tb}}} r_{\text{tb},\theta}^2(x_i)
\end{align}

Where\textbf{ $N_{int}$} are the number of collocation points or the PDE points, $N_{sb}$ are the points are each spatial boundary or boundary condition points and $N_{tb}$ are temporal boundary points or the Initial condition points.

Finally, we train our neural network to minimize the above loss terms and find the parameter $\theta$. 

\begin{equation}
\theta^\ast = \text{argmin}_{\theta} \Big(L_{F}(\theta) + L_B(\theta) + \lambda_I L_I(\theta)\Big)
\end{equation}

The weight ($\lambda_I$) for the temporal loss term is given by the equation:

\begin{equation}
    \lambda_I = C_t \left(1 - \frac{t}{T_{\text{max}}} \right) + 1
\end{equation}

where:
- \(\lambda_I\) is the weight for the temporal loss term,
- \(C_t\) is a constant,
- \(t\) is the current time,
- \(T_{\text{max}}\) is the maximum time.

In the first experiments, we added an approximation of the wave for c=1. We know the exact solution for that case, which is :
\begin{equation}
u(x, t) = \sin(x) \cdot (\sin(t) + \cos(t))
\end{equation}

For this experiment, we used a fully connected neural network comprising of 5 layers and 64 units. Sobol sequences were used to generate collocation points, spatial boundary points and temporal boundary points. Specifically, we generated 512 collocation points, 32 temporal points, and 64 points on each boundary. The network was optimized using an L-BFGS optimizer.

\begin{figure}[ht]
  \includegraphics[width=15cm]{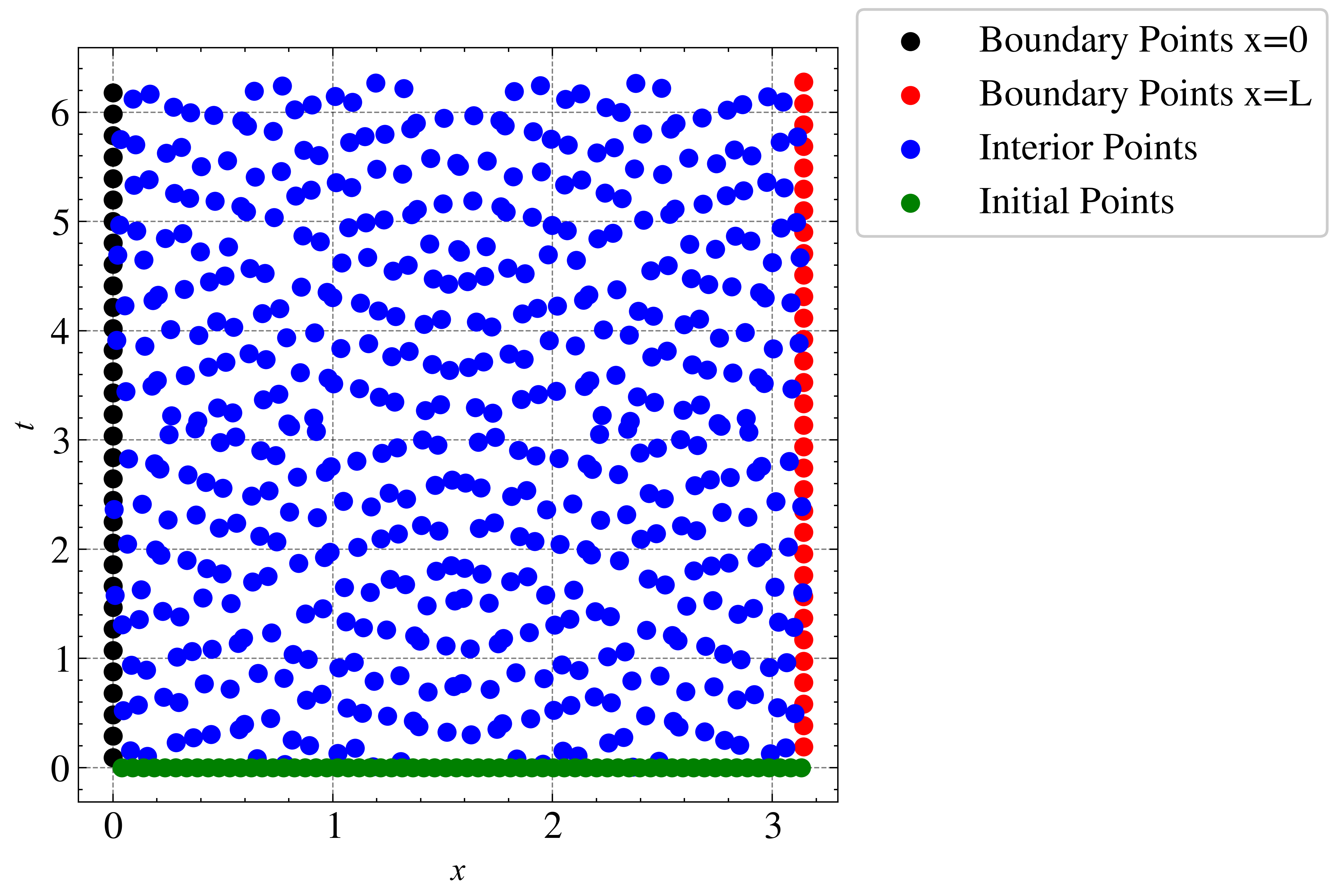} 
  \caption{Sampled points over the spatial and temporal domain. At \(x\) = 0 and \(x\) = L boundary points are added, which will define the value which u(x,t) will take.}
  \label{fig:geometry}  % Optional label for cross-referencing
\end{figure}

\subsubsection{Results for 1D Wave Equation}

\begin{figure}[H]
    \centering

    \begin{subfigure}{0.3\textwidth}
        \centering
        \includegraphics[width=1\linewidth]{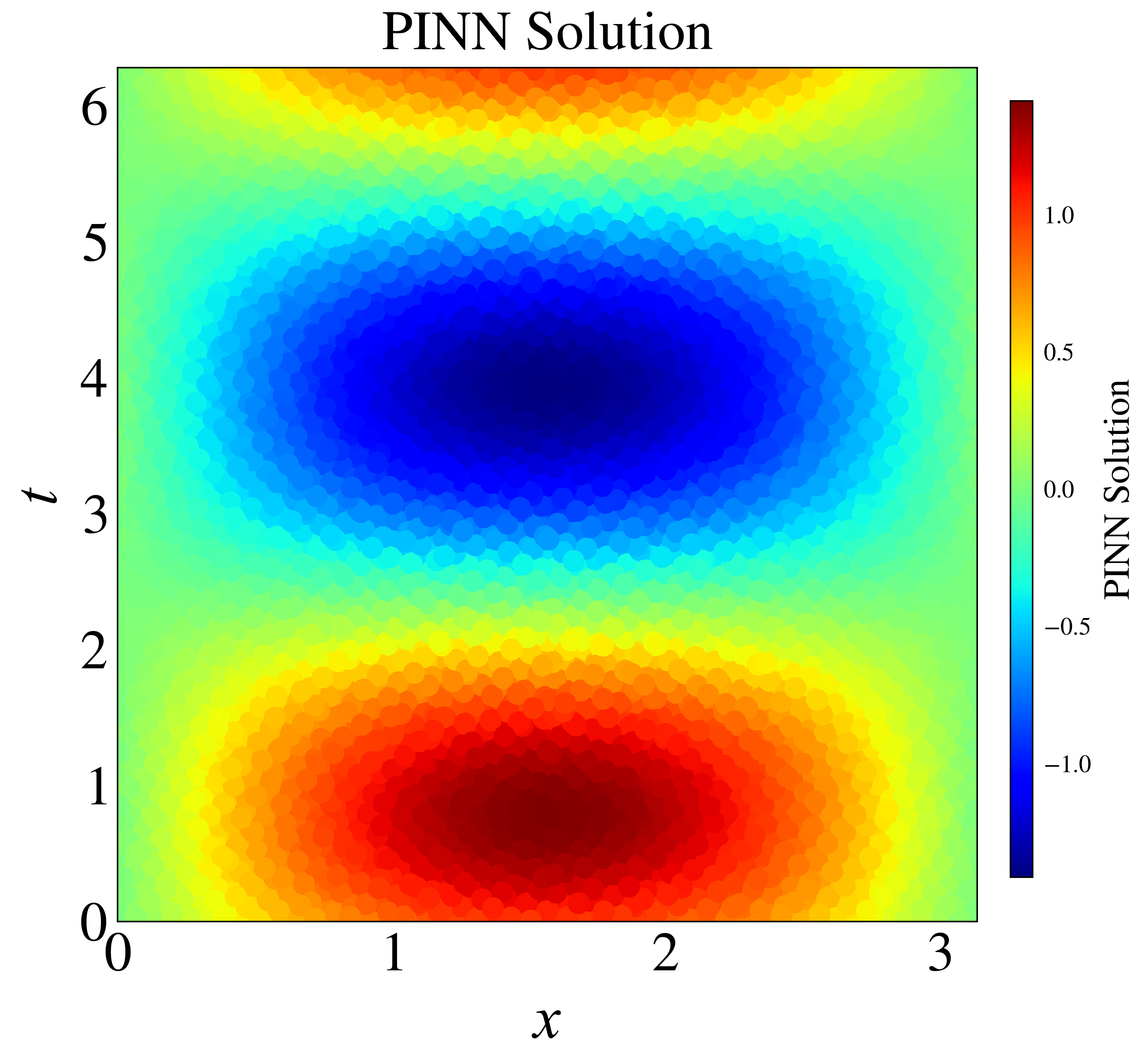}
        \caption{PINN Solution}
        \label{subfig_c1_1d_wave_pinn}
    \end{subfigure}
    \hfill
    \begin{subfigure}{0.3\textwidth}
        \centering
        \includegraphics[width=1\linewidth]{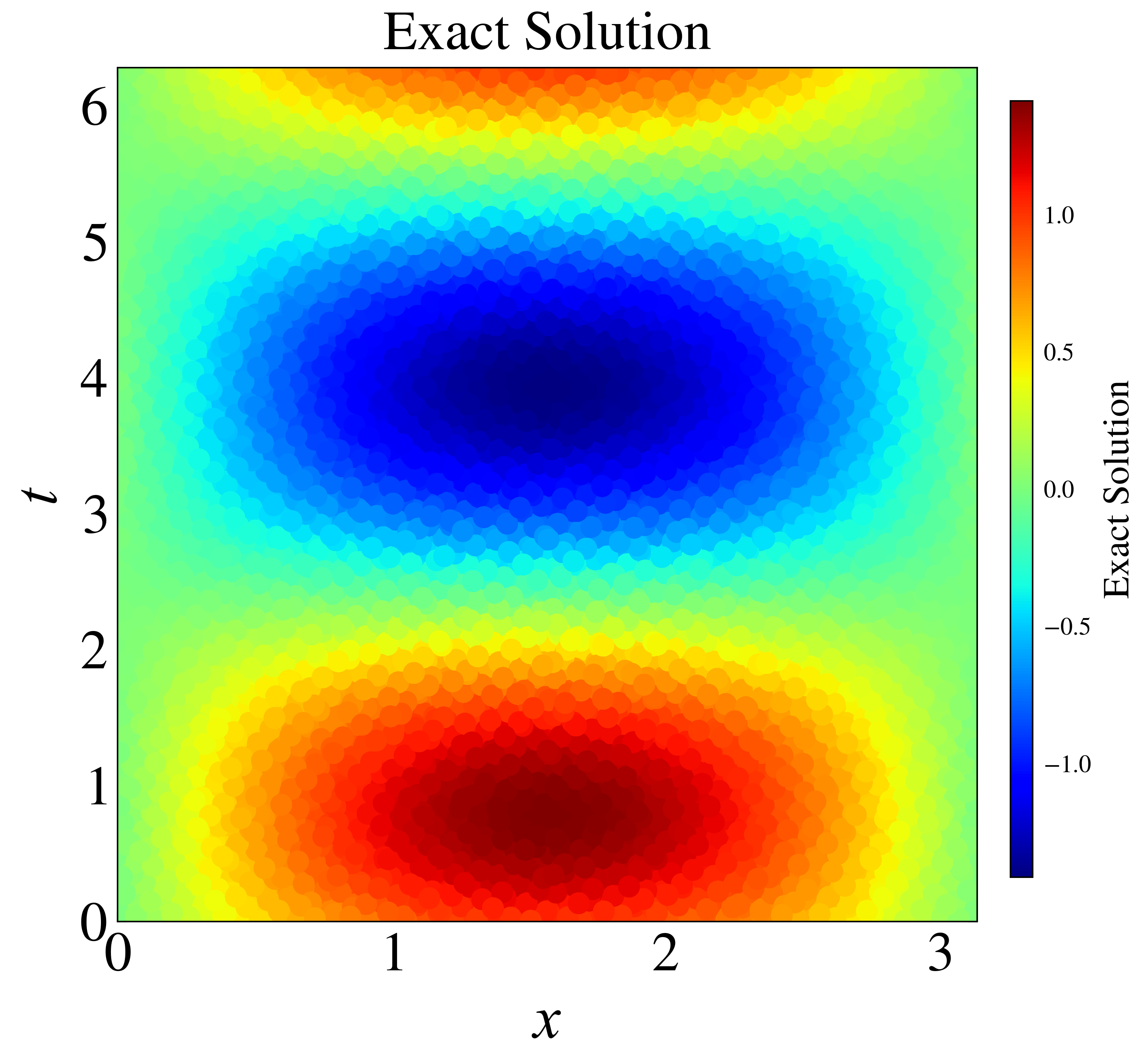}
        \caption{Exact Solution}
        \label{subfig_c1_1d_wave_exact}
    \end{subfigure}
    \hfill
    \begin{subfigure}{0.3\textwidth}
        \centering
        \includegraphics[width=1\linewidth]{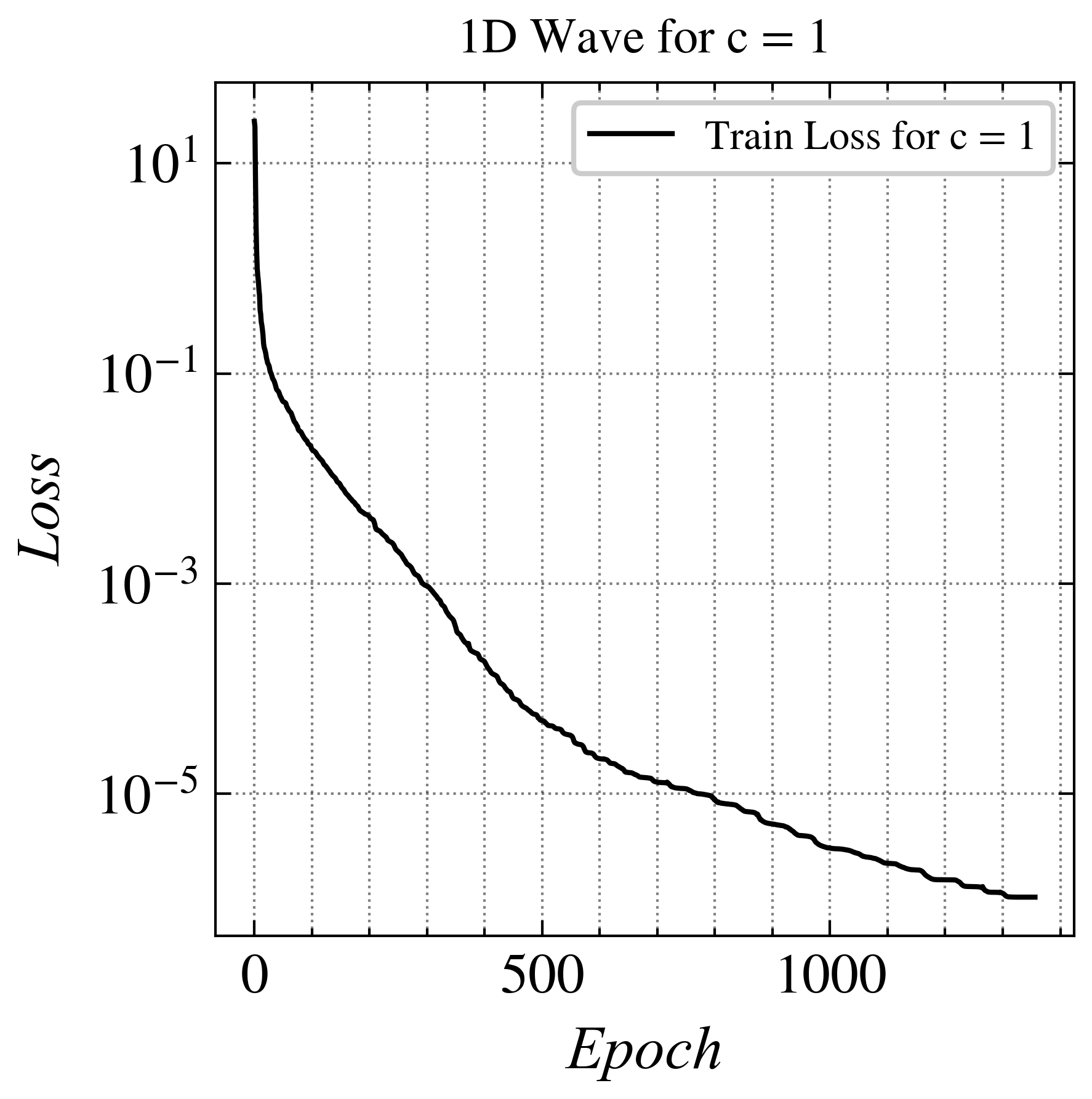}
        \caption{Loss }
        \label{subfig_c1_1d_wave_loss}
    \end{subfigure}
    
    \vspace{0.5cm}  % Adjust the vertical space between the first and second rows of subfigures

    \begin{subfigure}{0.3\textwidth}
        \centering
        \includegraphics[width=1\linewidth]{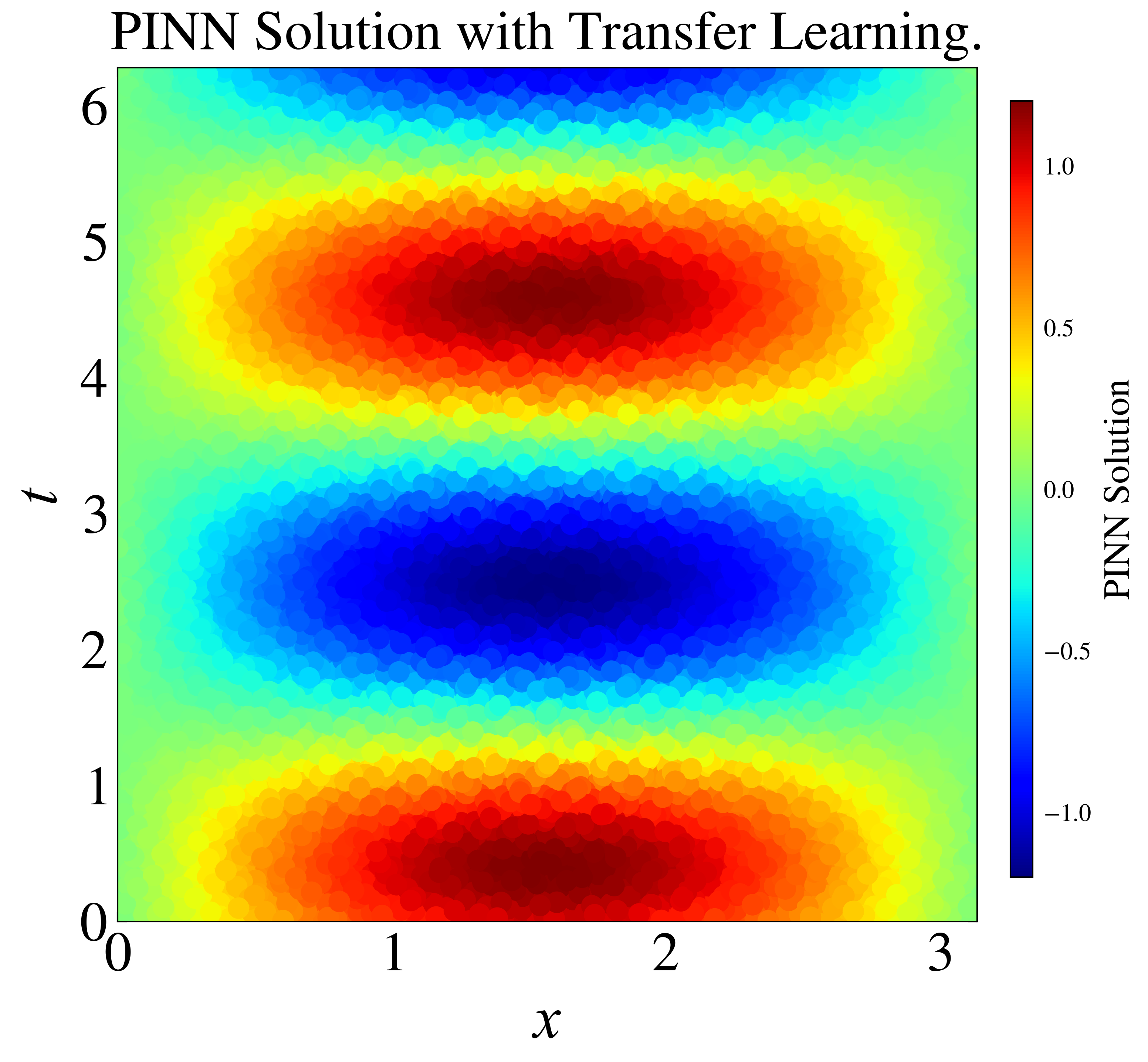}
        \caption{$c = 1.5$}
        \label{c_1_5_without_tl}
    \end{subfigure}
    \hfill
    \begin{subfigure}{0.3\textwidth}
        \centering
        \includegraphics[width=1\linewidth]{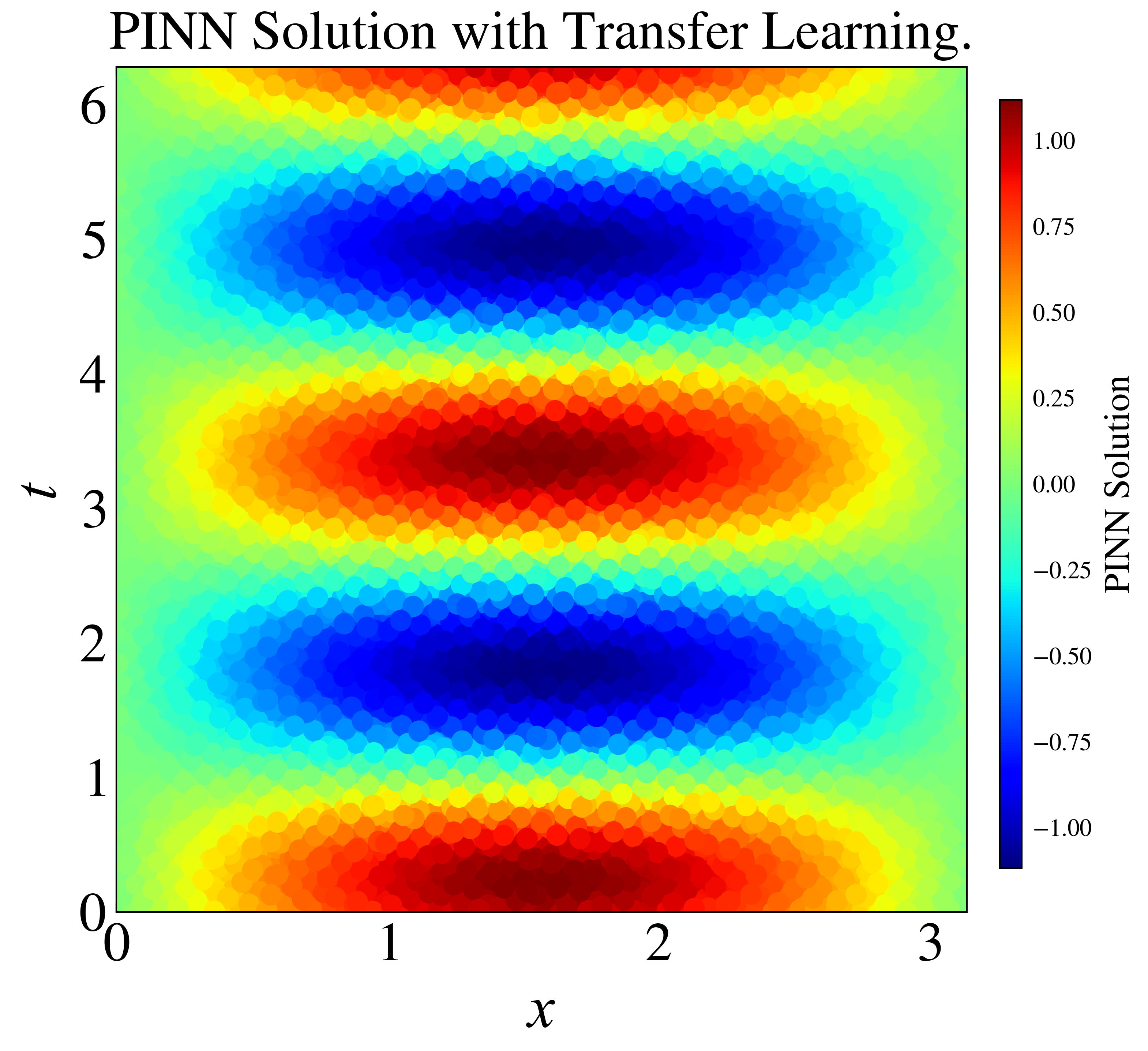}
        \caption{$c = 2$}
        \label{c_1_5_with_tl}
    \end{subfigure}
    \hfill
    \begin{subfigure}{0.3\textwidth}
        \centering
        \includegraphics[width=1\linewidth]{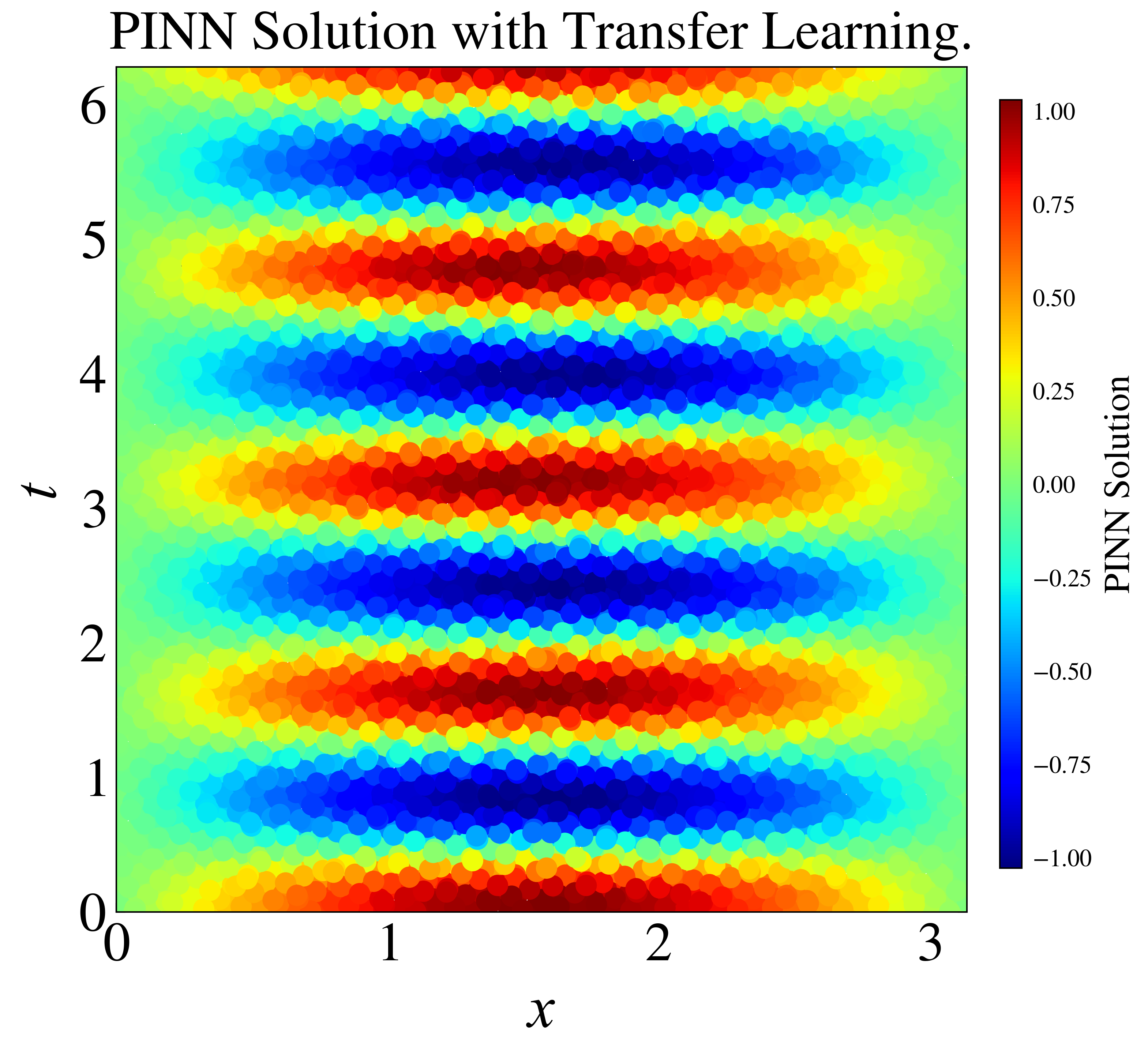}
        \caption{$c = 4$}
        \label{c_4_with_tl}
    \end{subfigure}

    \vspace{0.5cm}  % Adjust the vertical space between the second and third rows of subfigures

    \begin{subfigure}{0.3\textwidth}
        \centering
        \includegraphics[width=1\linewidth]{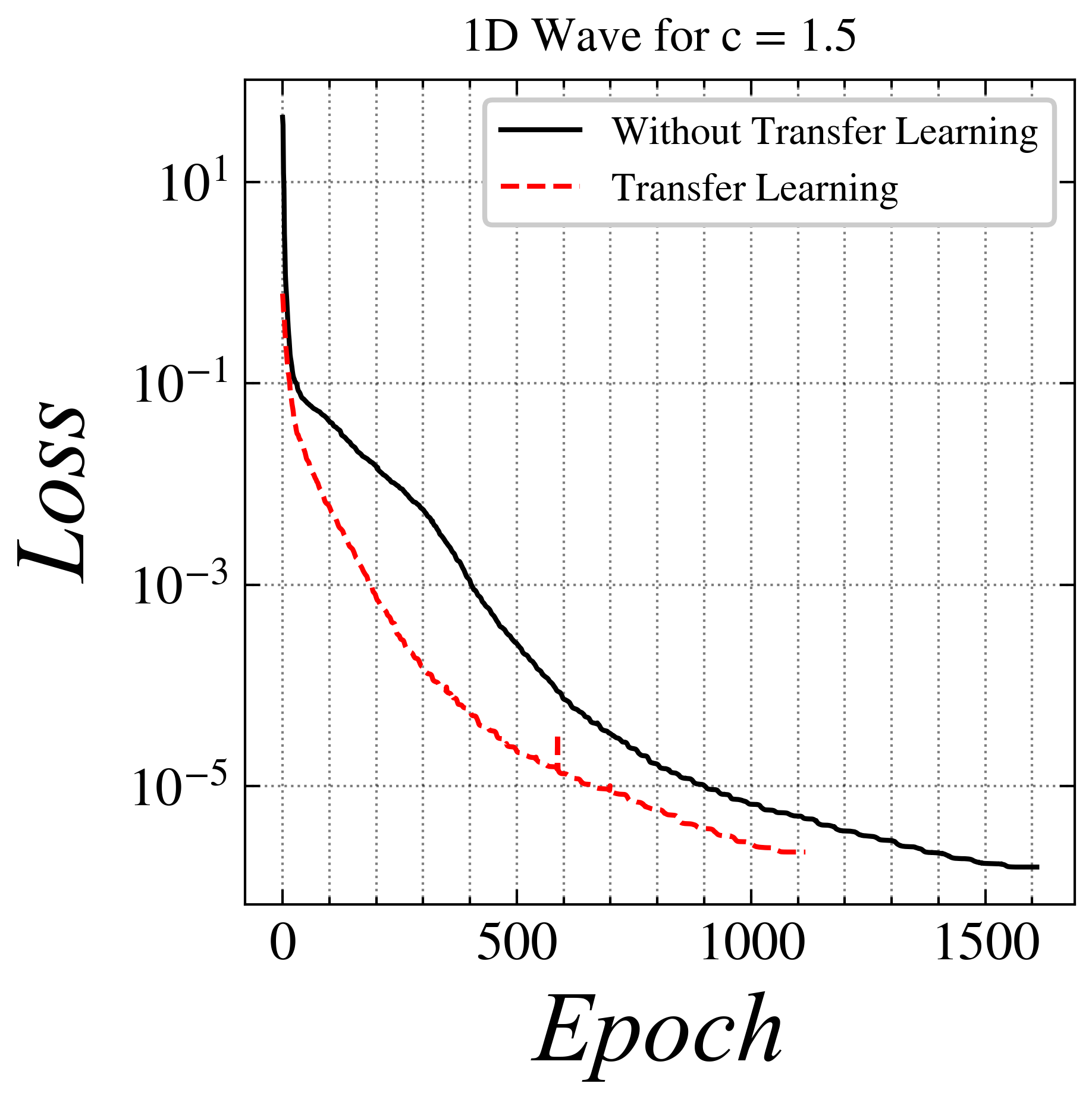}
        \caption{$c = 1.5$}
        \label{1d_c_1_5_compare}
    \end{subfigure}
    \hfill
    \begin{subfigure}{0.3\textwidth}
        \centering
        \includegraphics[width=1\linewidth]{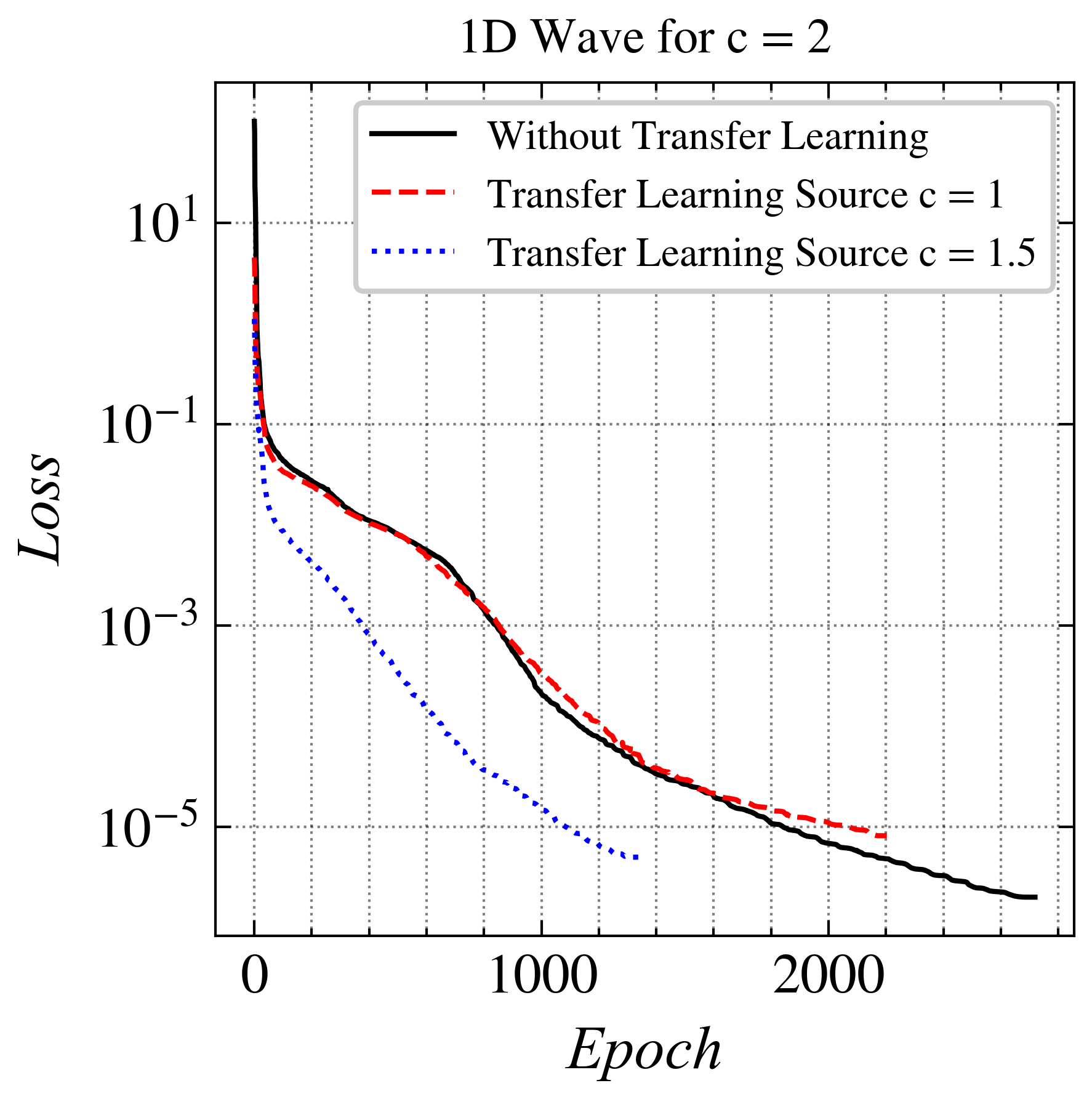}
        \caption{$c = 2$}
        \label{1d_c_2_loss}
    \end{subfigure}
    \hfill
    \begin{subfigure}{0.3\textwidth}
        \centering
        \includegraphics[width=1\linewidth]{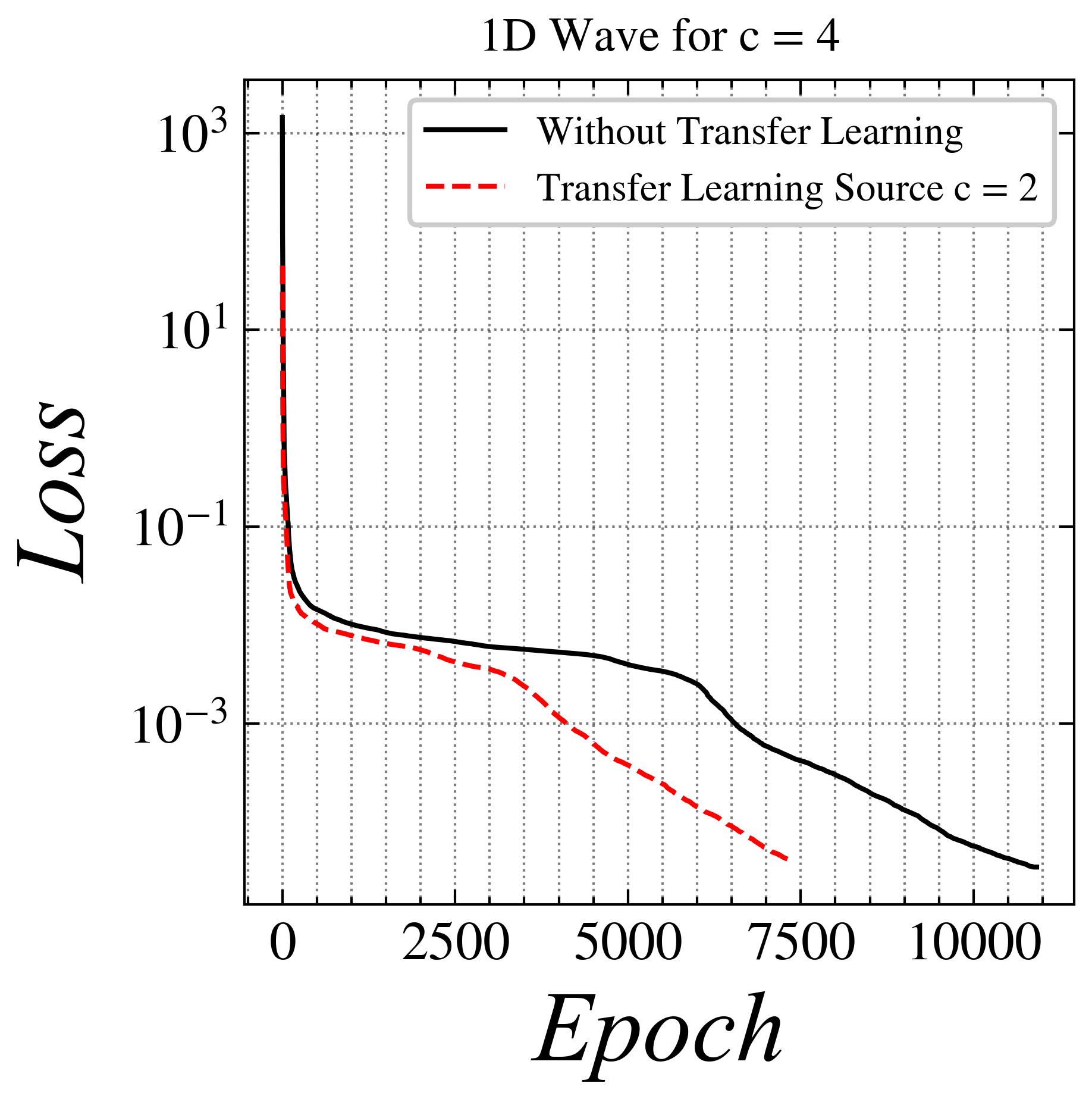}
        \caption{$c = 4$}
        \label{1d_c_4_loss}
    \end{subfigure}

    \caption{Results for $c = 1$ (L-BFGS): (a) Approximated solution by PINN, (b) Exact solution, (c) Loss  Results for various values of $c$ with transfer learning: (d) $c = 1.5$, (e) $c = 2$, (f) $c = 4$.}
    \label{fig_c_1_1d_wave}
\end{figure}

For c=1, the L2 Relative Error Norm between the exact solution and PINN solution is 0.05\%. 

Result \ref{subfig_c1_1d_wave_pinn} is the baseline or the source model that we trained, for c = 1. Now, in the upcoming results, we use this model as a source; as we increase the value of c to 1.5, 2.0, and 4.0. In figure [\ref{fig_c_1_1d_wave}], the model that used transfer learning performed better than the models without transfer learning. In \ref{fig2}, the loss reached an order of $10^{-5}$ in 600 epochs with transfer learning whereas it took 1000 epochs without transfer learning. \\

It can be observed in figure [\ref{1d_c_4_loss}] that the model without transfer learning took more than 10,000 epochs to converge, whereas the model which used transfer learning took only 7,500 epochs to converge. The order of loss is not as low as the other results because of the complexity of the solution. 

\section{Conclusions}
This work aims to shed light on the common challenges encountered by PINN when applied to high-frequency and multi-scale problems. We explored the potential of transfer learning as a viable solution to these problems. 
In the experiments, we observed that the PINN depicts the ability to approximate the harmonic oscillator at a frequency of 20 Hz. However, as the frequency increases, a noticeable increase in computational cost follows, accompanied by increased convergence times. 

The application of the vanilla PINN, utilizing an identical neural network architecture as the 20 Hz case, proves unfeasible in achieving convergence at 40 Hz, 50 Hz, and 60 Hz with the same amount of collocation points. While the model performs well on low-frequency problems, it starts struggling when given higher frequencies. 
Through transfer learning, we were able to learn the 50 Hz and 60 Hz solutions, without adding more layers or changing the number of collocation points. The results were promising as well, with a loss reaching an order of $10^{-2}$.

Similarly, in the context of the one-dimensional wave equation, with the use of transfer learning, we learned the PINN solution for different wave velocities, starting from 2 all the way up to 4.  The transfer learning method turned out to be effective, as for higher wave velocity, the model achieved convergence significantly quicker.

Transfer learning has proved to be an effective method for enhancing the efficiency and convergence characteristics of PINNs, preventing the necessity for modifications to the network architecture, which in turn causes more parameters. 
Future research will focus on exploring transfer learning methodologies in more complex scenarios, including the two-dimensional wave equation with different source terms.

\bibliographystyle{plain}
\bibliography{references}  %%% Uncomment this line and comment out the ``thebibliography'' section below to use the external .bib file (using bibtex) .

%%% Uncomment this section and comment out the \bibliography{references} line above to use inline references.
% \begin{thebibliography}{1}

% 	\bibitem{kour2014real}
% 	George Kour and Raid Saabne.
% 	\newblock Real-time segmentation of on-line handwritten arabic script.
% 	\newblock In {\em Frontiers in Handwriting Recognition (ICFHR), 2014 14th
% 			International Conference on}, pages 417--422. IEEE, 2014.

% 	\bibitem{kour2014fast}
% 	George Kour and Raid Saabne.
% 	\newblock Fast classification of handwritten on-line arabic characters.
% 	\newblock In {\em Soft Computing and Pattern Recognition (SoCPaR), 2014 6th
% 			International Conference of}, pages 312--318. IEEE, 2014.

% 	\bibitem{hadash2018estimate}
% 	Guy Hadash, Einat Kermany, Boaz Carmeli, Ofer Lavi, George Kour, and Alon
% 	Jacovi.
% 	\newblock Estimate and replace: A novel approach to integrating deep neural
% 	networks with existing applications.
% 	\newblock {\em arXiv preprint arXiv:1804.09028}, 2018.

% \end{thebibliography}

\end{document}